\RequirePackage{fix-cm}

\documentclass[smallextended]{svjour3}       
\smartqed  
\usepackage{amssymb}
\usepackage{graphicx}
\usepackage{amsmath}
\usepackage{mathtools}
\usepackage{booktabs}
\usepackage{multirow}
\usepackage[colorlinks=true, allcolors=blue]{hyperref}
\usepackage{adjustbox}
\usepackage{subfigure}
\def\hl{\setlength{\fboxsep}{1.0pt}\colorbox[rgb]{0.85,0.85,0.85}}
\usepackage{float}
\usepackage{lipsum}
\usepackage{ifmtarg}
\usepackage{array}
\usepackage{tabularx}
\graphicspath{{figures/}}

\usepackage{color}
\begin{document}

\title{A Review of Artificial Fish Swarm Algorithms: Recent Advances and Applications
}


\author{Farhad Pourpanah \and
        Ran Wang \and
        Chee Peng Lim\and
        Xi-Zhao Wang \and
        Danial Yazdani
}


\institute{Farhad Pourpanah \at
              College of Mathematics and Statistics, Guangdong Key Lab. of intelligent Information Processing, Shenzhen University, China.
              \at Department of Electrical and Computer Engineering, University of Windsor.\\
              \email{farhad.086@gmail.com}           
           \and
           Ran Wang \at
           College of Mathematics and Statistics, Shenzhen Key Lab. of Advanced Machine Learning and Applications, Guangdong Key Lab. of intelligent Information Processing, Shenzhen University, China.\\
              \email{wangran@szu.edu.cn}   
           \and
           Chee Peng Lim \at
           Institute for Intelligent Systems Research and Innovation, Deakin University, Australia.\\
              \email{chee.lim@deakin.edu.au}
           \and
           Xi-Zhao Wang \at
           College of Computer Science and Software Engineering, Guangdong Key Lab. of intelligent Information Processing, Shenzhen University, China.\\
              \email{xizhaowang@ieee.org}
          \and
           Danial Yazdani \at
           School of Computer Science and Engineering, Southern University of Science and Technology, China.\\
              \email{danial.yazdani@yahoo.com}
}

\date{Received: date / Accepted: date}

\maketitle

\begin{abstract}
{\color{black}
The Artificial Fish Swarm Algorithm (AFSA) is inspired by the ecological behaviors of fish schooling in nature, viz., the preying, swarming and following behaviors. Owing to a number of salient properties, which include flexibility, fast convergence, and insensitivity to the initial parameter settings, the family of AFSA has emerged as an effective Swarm Intelligence (SI) methodology that has been widely applied to solve real-world optimization problems. Since its introduction in 2002, many improved and hybrid AFSA models have been developed to tackle continuous, binary, and combinatorial optimization problems. This paper aims to present a concise review of the continuous AFSA, encompassing the original ASFA, its improvements and hybrid models, as well as their associated applications.  We focus on articles published in high-quality journals since 2013. Our review provides insights into AFSA parameters modifications, procedure and sub-functions.  The main reasons for these enhancements and the comparison results with other hybrid methods are discussed.  In addition, hybrid, multi-objective and dynamic AFSA models that have been proposed to solve continuous optimization problems are elucidated.  We also analyse possible AFSA enhancements and highlight future research directions for advancing AFSA-based models.}


\end{abstract}

\section{Introduction}
\label{Sec:intro}
As part of artificial intelligence, the Swarm Intelligence (SI)~\cite{blum2008swarm,pourpanah2019feature} methodology is largely inspired by the collective behaviors of biological and natural evolutional phenomena, e.g. swarms of worms, bees and ants, schools of fish and flock of birds.
The main reasons for the recent popularity of SI-based algorithms in solving optimization problems include their capability of self-learning, fast convergence, flexibility, simple structure, insensitivity to initial parameters and adaptability to external variants~\cite{fister2013comprehensive,li2021diversity}.
They have been applied to undertake various problems successfully, such as image processing~\cite{rifaei2015deploying,pajouhi2018image}, feature selection~\cite{gholami2018feature,pourpanah2016ahybrid,pourpanah2019feature1}, medical diagnosis~\cite{alkeshuosh2017using,zomorodi2019hybrid} and training machine learning models~\cite{koohestani2019integration,pourpanah2018anomaly}, dynamic optimization problems~\cite{jia2019animproved,pourpanah2019mbso}.
In general, SI-based models start with a set of random solutions.
Each member of swarm evolves individually, and is guided toward a better position iteratively in the search space until the stopping criterion is satisfied~\cite{pourpanah2019ahybrid,pourpanah2017aqlearning,pourpanah2018non}.
The evolving behavior towards optimality is usually achieved by the self-organizing capability of the swarm through simple interaction rules.
The popular SI-based optimization algorithms include particle swarm optimization (PSO)~\cite{kennedy2010particle}, ant colony optimization (ACO)~\cite{dorigo1997ant}, firefly algorithm (FA)~\cite{fister2013comprehensive,yang2010nature}, grey wolf optimizer (GWO)~\cite{mirjalili2014grey}, crow search algorithm (CSA)~\cite{askarzadeh2014anovel}, artificial fish swarm algorithm (AFSA)~\cite{Li2002optimizing}, cuckoo search (CS)~\cite{yang2009cuckoo}, bat algorithm (BA)~\cite{yang2010new} and brain storm optimization (BSO)~\cite{shi2011brain}.\par

Comparatively, AFSA~\cite{Li2002optimizing} is inspired by the behaviors of a fish swarm, which is a more recent SI-based optimization method introduced by Li in 2002.
It is a stochastic algorithm that searches for a set of solutions in a randomized procedure to solve NP-hard problems.
The basic idea is to simulate a number of ecological behaviors of fish schooling in the water.
The first behavior is praying or foraging, where every fish in nature probes for its prey individually within its visual distance.
The second behavior is swarming or clustering, where every fish tends to avoid danger by assembling in a group.
The third behavior is the following or rear-end, where once a fish finds a food source with a high concentration, other fishes in its neighborhood tend to reach the food.
The AFSA considers each food position as a feasible solution for the optimization problem in hand, and the density of the corresponding food indicates the quality of the solution that can be measured by a fitness function.
The AFSA uses the preying behavior to identify the optimal solution, the following behavior to escape from the local optimal solutions, and the swarming behavior to gather the fish swarm around the optimal solution.\par 

\begin{figure}[tb!]
\centering
\includegraphics[width=0.9\linewidth]{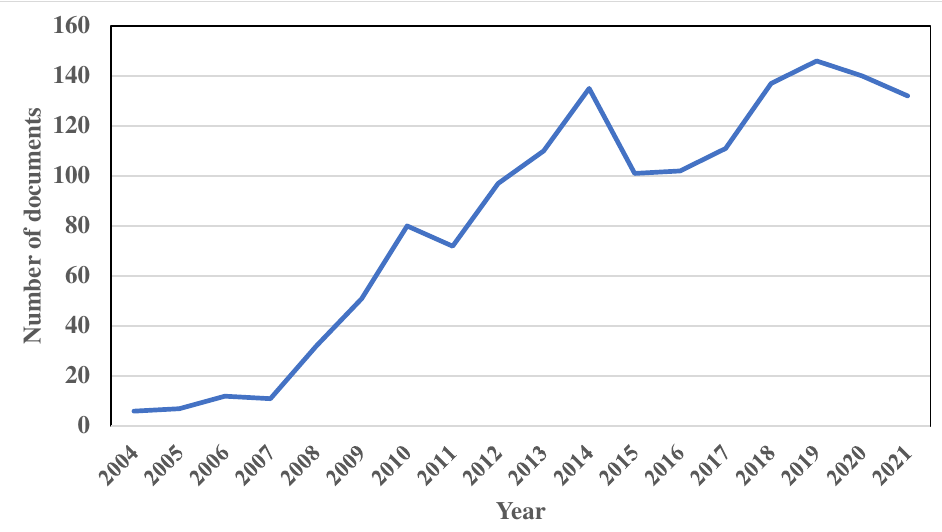}
\caption{ \color{black}The publication trend of AFSA from 2004 to 2021. The result obtained by conducting searches based on the exact phrases of ``Artificial Fish Swarm Algorithm", ``Artificial Fish Swarm Optimization", ``Artificial Fish Schooling Algorithm" in Title, Abstract, or Keywords of the documents. Source: {\color{black}https://www.scopus.com} }
\label{Fig:statistic}
\end{figure}

Since its introduction in 2002, many improved and hybrid models of AFSA have been developed to solve real-world optimization problems, including data clustering~\cite{yazdani2013new,feng2020analysis}, image segmentation~\cite{lei2018image}, fault diagnosis~\cite{zhu2017adaptive}, power allocation scheme~\cite{zhou2018artificial}, parameter optimization of the deep auto-encoder~\cite{shao2017novel}, spoiled meet detection~\cite{chen2018application}, manufacturing~\cite{maji2018optimal}, risk probability prediction~\cite{li2019risk}, guidance error estimation~\cite{zhou2019guidance}, navigation~\cite{Du2018Precision}, wireless sensor network~\cite{mechta2017prolonging,gorgich2021proposing}, and energy management~\cite{Talha2018Energy}.
Fig.~\ref{Fig:statistic} shows the publication trend of AFSA from 2004 to {\color{black}early} 2021. 
As can be seen, there is an upward trend in using AFSA.
A survey on the AFSA from its introduction to 2012 can be found in~\cite{neshat2014artificial}.
However, more than {\color{black}1000} articles, conference papers, and book chapters have been published since 2013, which are yet to be reviewed.\par

In this study, we aim to review the AFSA modifications that have been proposed to solve continuous optimization problems from 2013 onwards.
We mainly focus on articles published in key indexed journals by conducting searches in ScienceDirect, Springer, IEEE Xplore, and Google Scholar.
The keywords include ``artificial fish swarm algorithm", ``fish schooling", ``AFSA", ``AFs", ``hybrid AFSA", ``dynamic AFSA", ``multi-objective AFSA" and ``swarm intelligence".
In addition, articles that report the original AFSA without modifications are excluded, in order to focus the review on recent advances pertaining to the AFSA.
The scope of our review is focused on the AFSA modifications, specifically parameters modifications, procedure and sub-functions, hybrid, multi-objective and dynamic AFSA that have been proposed to solve continuous optimization problems.

Based on this taxonomy, this review consists of six sections.
Section~\ref{sec:background} explains various continuous optimization problems, the AFSA procedure and the three associated behaviors in detail, and compares the AFSA with PSO.
Section~\ref{Sec:taxo} provides a taxonomy of the AFSA modifications and reviews each modification.
Section~\ref{Sec:app} discusses the real-world applications of the AFSA.
Section~\ref{Sec:future} {\color{black} first empirically compares AFSA and its variants with several population based algorithms, and then} provides several suggestions for future research.
The concluding remarks are given in Section~\ref{Sec:con}.

\section{Background}
\label{sec:background}
In this section, we firstly formulate various continuous optimization problems including static optimization, static multi-objective optimization and dynamic optimization problems.
Then, the structure of the original AFSA along with its behaviours are described in detail. 
A comparison between the AFSA and PSO, which is the most popular SI algorithm, is also provided.  
\subsection{Optimization Problems}
\subsubsection{Static Optimization Problem}

In general, a static optimization problem for solving the minimum problem can be defined as follows:
Let $\vec{x}=(x_1,x_2,\ldots,x_d)$ indicates a vector of $d$ decision variables in the space $X$, and $f: X\to R$ is the objective function. 
The goal (for a minimum problem) is to find an optimal solution $\vec{x}\in X$ that minimizes the objective function value $f$, as follows: 
\begin{align}
    \min_{\vec{x}\in X} f(\vec{x}).
\end{align}

\subsubsection{Multi-Objective Optimization Problem}
A multi-objective optimization problem (MOOP) consists of $K$ objective functions that need to be optimized, as follows:
\begin{align}
\min_{\vec{x}\in X} & f_k(\vec{x}), \; k=1,2,\ldots,\hat{k},
\end{align}
where $f_k(\vec{x})$ represents the $k$-th objective function.

\subsubsection{Dynamic Optimization Problem}

A dynamic optimization problem (DOP) can be defined as: 
\begin{align}
\label{eq:DOP1}
 \min_{\vec{x}\in X} f\left(\vec{x},\vec{\alpha}^{(t)}\right),
\end{align}
where $\vec{\alpha}$ is a vector of time-varying control parameters with respect to the objective function, and $t\in [0,T]$ is the time index. 
Most existing works in the DOP literature consider the DOPs whose  environmental changes happen only in discrete time, i.e., $t \in \{1,\dots,T\}$.
In this respect, for a DOP with $T$ environmental states, there is a sequence of $T$ stationary environments:
\begin{align}
\label{eq:DOP2}
\left\langle	f(\vec{x},\vec{\alpha}^{(1)}),f(\vec{x},\vec{\alpha}^{(2)}), \dots ,f(\vec{x},\vec{\alpha}^{(T)})\right\rangle.
\end{align}

\subsubsection{Constrained Optimization Problem}
All aforementioned optimization problems can be subjected to constraints:
\begin{align}
    g_j(\vec{x})\geq 0, \; j = 1, \ldots, \hat{j},\\
    h_l(\vec{x})= 0, \; l = 1, \ldots, \hat{l},
\end{align}
where $g_j$ and $h_l$ are inequality and equality constraints, respectively, which define the feasibility of solutions.

\subsection{Artificial Fish Swarm Algorithm}
\label{Sec:AFSAa}
\begin{figure}[tb]
 \begin{center}
 \includegraphics[scale=0.70]{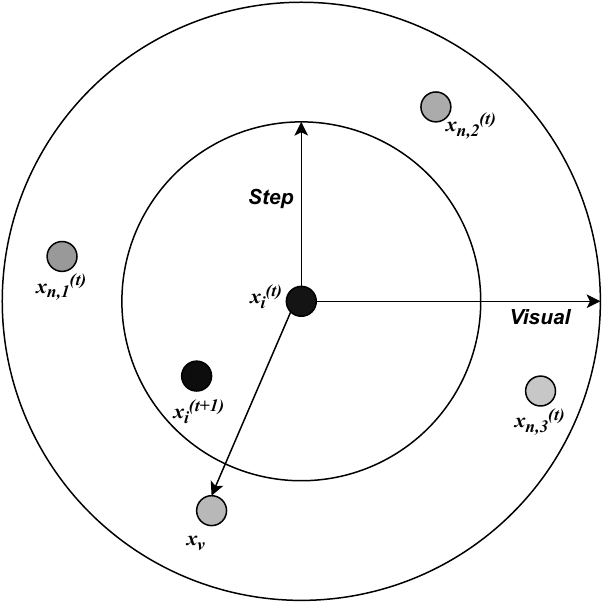}
  \end{center}
  \vspace*{-0.5cm}
  \caption{An artificial fish and its environment. Where {$x_i^{(t)}$} and \textbf{$x_i^{(t+1)}$} represent the current and next positions of the \textit{i}-th AF, respectively, ${x_v}$ is a random position inside the line of sight, and $x_{n,1}^{(t)}$, $x_{n,2}^{(t)}$ and $x_{n,3}^{(t)}$ are AFs which are located inside the field of vision of the \textit{i}-th AF. }
  \label{Fig:af}
\end{figure}

In AFSA, candidate solutions are called artificial fish (AF).
Fig.~\ref{Fig:af} shows a schema of an AF and its surrounding environment, where $\vec{x}=(x_1,x_2, \ldots,x_d)$ is an AF in a $d$-dimensional continuous space, \emph{Step} ($s$) shows the maximum Euclidean distance that an AF can relocate at each step, \emph{Visual} ($v$) is the maximum length of vision (Euclidean distance) of an AF, and $\vec{x}_v$ denotes a position inside the line of sight of the AF, $\vec{x}_{n,i}$ is a neighbor of the $n$-th AF located within the line of sight of the $i$-th AF, i.e., their Euclidean distance is less than $v$.
Each AF searches inside a hyper-ball whose center and radius are the AF position and $v$, respectively.
If an AF finds a position inside the hyper-ball that has a better fitness value than its current position, the AF moves a step toward it.

The AFSA starts by initializing $n$ randomly distributed AFs across the search space, where $n$ is the population size.
Each AF searches for positions with higher fitness values using three \emph{behaviors}: preying, swarming, and following.
The AFSA uses a \emph{bulletin board} to record the best found position so far.
At the end of each iteration, the best AF (i.e., the AF with the best fitness value) is compared with the archived solution in the bulletin board, and if the new AF is better, the bulletin's position will be updated.
This search cycle continues until the stop criterion is met.

\subsubsection{Behavior Descriptions}
\label{Sec:Sec:beh}
The three possible behaviors of an AF are described in the following.
\paragraph{Preying behavior}
\label{Sec:Sec:Sec:prey}
Preying is the basic behavior for a fish to move to a location with a higher concentration of food.
In AFSA, this behavior is modeled based on the current position of an AF and its surrounding area based on its field of view.
The $i$-th AF searches inside its field of vision randomly by:
\begin{align}
\label{eq:prey1}
\vec{x}_v=\vec{x}_i^{(t)}+ \left(v \cdot \mathcal{U}(0,1) \cdot \frac{\vec{u}}{\|\vec{u}\|}\right),
\end{align}
where $\vec{x}_i^{(t)}$ is the current position of the $i$-th AF, $\vec{x}_v$ is a random position inside the field of vision of the $i$th AF, $v$ is the visual length, $\mathcal{U}(0,1)$ is a random number drawn from a uniform distribution in (0,1), $\vec{u}$ is a vector of uniformly distributed numbers in $(-1,1)$, $\|\vec{u}\|$ is the $L2$-norm (Euclidean length) of $\vec{u}$, and $\frac{\vec{u}}{\|\vec{u}\|}$ provides a unit vector.

If $f(\vec{x}_v) \leq f(\vec{x}_i^{(t)})$ (in minimization problems), the $i$-th AF moves a step toward $\vec{x}_v$ using the following formula:
\begin{align}
\label{eq:prey2}
\vec{x}_{p,i}=\vec{x}_i^{(t)}+ \left(s \cdot \mathcal{U}(0,1) \cdot \frac{\vec{x}_v-\vec{x}_i^{(t)}}{\|\vec{x}_v-\vec{x}_i^{(t)}\|}\right),
\end{align}
where $\vec{x}_{p,i}$ is the position of the $i$-th AF after executing praying behaviour, which is a position alongside the $\vec{x}_v-\vec{x}_i^{(t)}$, and $\|\vec{x}_{p,i}-\vec{x}_i^{(t)}\| \leq s$.
Otherwise, if $f(\vec{x}_v) > f(\vec{x}_i^{(t)})$, another position $\vec{x}_v$ is randomly selected using \eqref{eq:prey1}.
The aforementioned process is repeated until a position with a better fitness value is found. 
In the case where the $i$-th AF cannot found a better position after trying $\mathfrak{t}$ positions, which is called \emph{try-number}, it moves a random step:
\begin{align}
\label{eq:prey3}
\vec{x}_{p,i}=\vec{x}_i^{(t)}+ \left(v \cdot \mathcal{U}(0,1) \cdot \frac{\vec{u}}{\|\vec{u}\|}\right).
\end{align}

\paragraph{Swarming behavior}
\label{Sec:Sec:Sec:swarm}
In nature, a fish tends to assemble in a group to avoid danger, while avoids over-crowded areas.
This behavior of fish swarms is also modeled in AFSA.
Let  $\vec{x}_c$ be the central position of the swarm which is the average of the positions of all AFs:
\begin{align}
\label{eq:center}
\vec{x}_{c}= \frac{1}{n} \sum _{k=1}^{n} \vec{x}_{k}.
\end{align}

The $i$-th AF moves toward $\vec{x}_{c}$ if $f(\vec{x}_c) \leq f(\vec{x}_i)$ and the area around $\vec{x}_{c}$ is not over-crowded.
In AFSA, the over-crowding status around a position $\vec{x}$ is determined by:  
\begin{align}
\label{eq:swcon}
\mathbb{C}(\vec{x})=
\begin{dcases}
1, & \mathrm{if~~}\frac{\mathbb{N}(\vec{x})}{n}> \delta \\
0, & \mathrm{otherwise}
\end{dcases},
\end{align}
where $\mathbb{C}(\vec{x})=1$ indicates that the area around the $\vec{x}$ is overcrowded, $\mathbb{N}(\vec{x})$ represents the number of AFs whose Euclidean distances to the $i$-th AF is less than $v$, and $\delta \in [0,1]$ is the \emph{crowding factor}.
By avoiding movements toward over-crowded places, AFSA tries to maintain the swarm diversity to avoid collapsing AFs in a limited areas and premature convergence.
If $\mathbb{C}(\vec{x}_c)=0$ and $f(\vec{x}_c) \leq f(\vec{x}_i)$, the $i$-th AF moves a step toward the center position:
\begin{align}
\label{eq:swarm}
\vec{x}_{s,i}=\vec{x}_i^{(t)}+ \left(s \cdot \mathcal{U}(0,1) \cdot \frac{\vec{x}_c-\vec{x}_i^{(t)}}{\|\vec{x}_c-\vec{x}_i^{(t)}\|}\right),
\end{align}
otherwise, the $i$-th AF will execute the preying behavior.
\paragraph{Following behavior}
\label{Sec:Sec:Sec:follow}
When an AF finds a location with a better concentration of food, other fish will follow it.
For the $i$-th AF, if there is at least another AF (indicated by $j$) where $\|\vec{x}_j-\vec{x}_i\| \leq v$, then the $i$-th AF moves a step toward the $j$-th AF position by:
\begin{align}
\label{eq:follow}
\vec{x}_{f,i}=\vec{x}_i^{(t)}+ \left(s \cdot \mathcal{U}(0,1) \cdot \frac{\vec{x}_j-\vec{x}_i^{(t)}}{\|\vec{x}_j-\vec{x}_i^{(t)}\|}\right),
\end{align}
otherwise, the $i$-th AF will execute the preying behavior.


\subsubsection{The AFSA procedure}

\begin{figure}[htb!]
 \begin{center}
 \includegraphics[scale=0.9]{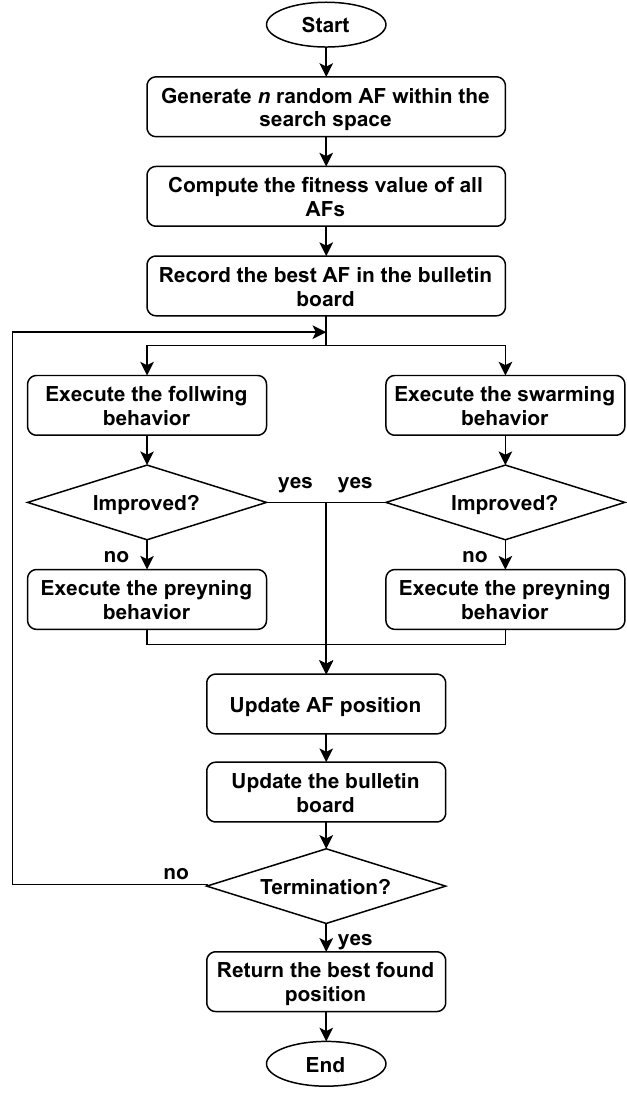}
  \end{center}
  \vspace*{-0.5cm}
  \caption{Flowchart of the original AFSA. }
  \label{Fig:flow}
\end{figure}

 The AFSA starts with generating $n$ random AFs within the search space.
Next, it simultaneously executes the swarming and following behaviors for each AF.
In the situation when the criteria to execute the behaviors are not satisfied, or no improvement is observed after executing these behaviors, the AFSA executes preying behavior.
Then, replacement takes place if any of the generated positions performs better than the best found position recorded in the bulletin board.  
This procedure continues until the stop condition is met.  
Fig.~\ref{Fig:flow} summarizes the AFSA procedure. 

\begin{figure}[tb!]
\centering
\includegraphics[width=0.65\linewidth]{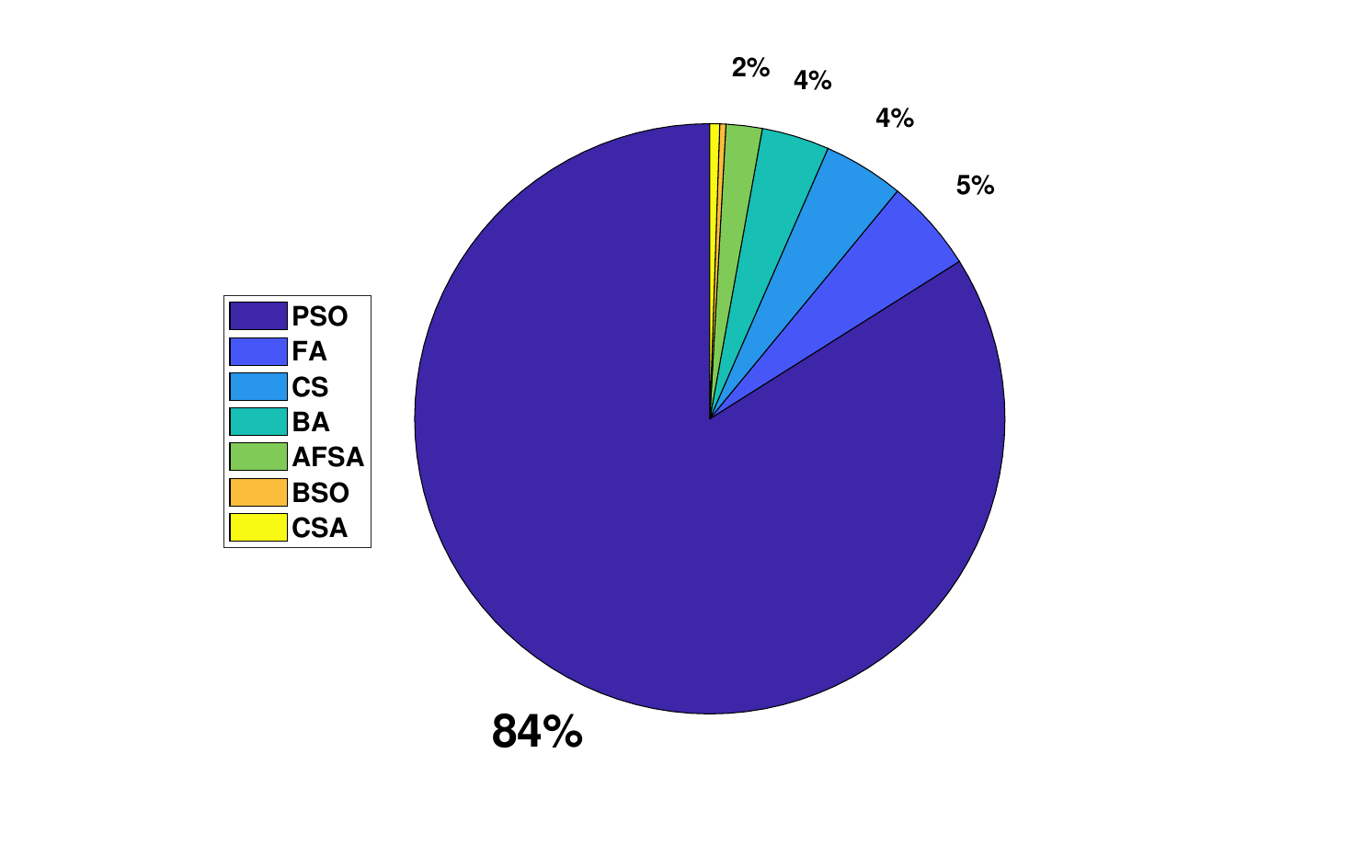}
\caption{{\color{black}The popularity of PSO as compared with other SI-based algorithms, i.e., FA, CS, BA, AFSA, BSO and CSA.} Source: {\color{black}https://www.scopus.com}}
\label{Fig:com}
\end{figure}

{\color{black}
\subsection{A short comparison between AFSA, PSO and FA}
The AFSA and its variants are useful for solving continuous optimization problems. They have demonstrated good results as compared with those from other population-based optimization models. A comparison among the AFSA, PSO and FA is made. 
PSO is one of the most popular SI algorithms. Fig.~\ref{Fig:com} shows the popularity of PSO as compared with other SI-based algorithms, which include FA, CS, BA, AFSA, BSO and CSA. The results are obtained through a search in title, abstract, or keywords of the associated publications (over 100,000 articles). It can be observed that PSO occupies 84\% of all the articles. While FA uses a similar mechanism as the AFSA, i.e., distance-based metrics, to determine its neighborhood fireflies. Specifically, several key differences from the structure, procedure, and update rules among these three algorithms can be observed, as follows:

\begin{itemize}
  \item In PSO, the candidate solutions are particles. Each particle keeps the information of its personal best ($\vec{\mathbf{p}}$) position found over the previous iterations, which is used as an attractor to determine the subsequent position. Comparatively, each AF in the AFSA and each firefly in the FA do not keep any previous information, except their current positions, in the memory structure.
  
  \item In PSO, the neighborhood particles are determined according to their indices. In the AFSA and FA, however, the neighborhood AFs and fireflies are determined according to the Euclidean distances.

  \item In all three algorithms, the candidate solutions are attracted toward the better neighbor candidate solution.  In PSO, the attractors are chosen among the personal best solutions of particles and the neighborhood topology. In the FA, fireflies are attracted to another (fitter) firefly, where the attractiveness of each firefly is proportional to its light intensity. In the AFSA, however, the attractors of each AF are the best AF in its visual range and the center position of the swarm.

  \item In the AFSA, the movement length is limited by parameter $s$, while there is no limitation in the movement size in the FA and PSO.  In PSO, the movement length is defined by the velocity vector, which can lead to a higher convergence speed as compared with that of the AFSA. Nevertheless, the trade-off is a possible pre-mature convergence in PSO. In contrast, in the FA, each firefly moves according to three factors, i.e., its current position, another attractive firefly and random walk.

  \item Both the FA and PSO rapidly loose the diversity of their populations, which can lead to an increased probability of being trapped in local optima. On the other hand, in the AFSA, the selection rules, which are defined based on $\delta$, do not allow the AFs to concentrate on an optimum location. This results in maintaining a higher degree of diversity over time and increasing the AFSA exploration capability. However, the trade-off is a reduction in the AFSA convergence speed.

  \item In both the FA and PSO, the social attractors and personal best information are unified in a formula, in which the next position is dependent on both individual and group attractors simultaneously. In contrast, in the AFSA, the update rules are separated as behaviors that are performed on the AFs according to certain conditions. Therefore, in each iteration, each AF can either perform local search around itself or a social-based behavior and move towards other attractors, therefore, enhancing its search capability.
 
\end{itemize}

}

\begin{figure}[tb!]
 \begin{center}
 \includegraphics[scale=0.850]{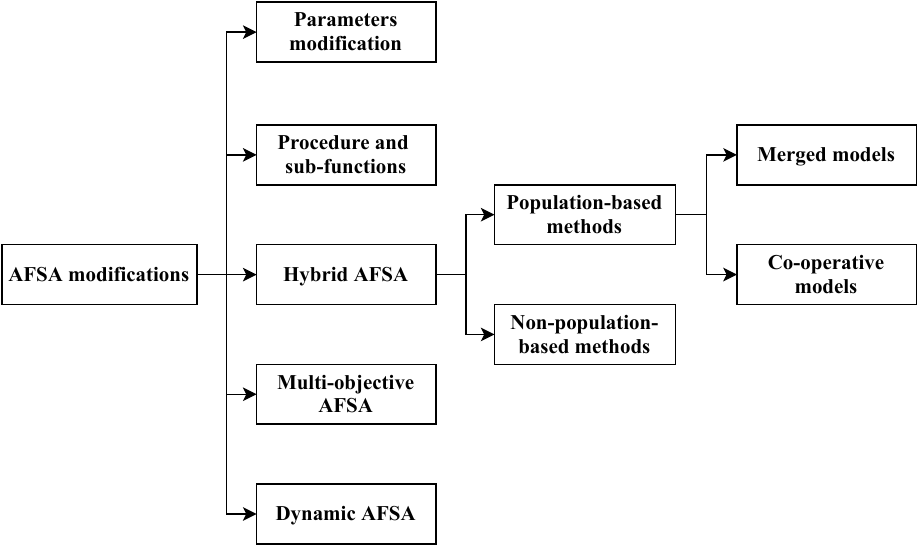}
  \end{center}
  \vspace*{-0.5cm}
  \caption{The taxonomy of the AFSA modifications.}
  \label{Fig:taxo}
\end{figure}

\section{A Taxonomy of the AFSA Modifications}
\label{Sec:taxo}

In this section, a taxonomy of the AFSA modifications is provided.
According to the taxonomy (Fig.~\ref{Fig:taxo}), these modifications are classified into the following categories.

\begin{itemize}

\item\textit{ Parameter control}:
This group of modifications focuses on controlling the parameter values of the AFSA, especially the visual distance ($v$), maximum step length ($s$) and the crowding factor ($\delta$), in order to control the trade-off between the exploration and exploitation capabilities of the AFSA.
To this end, different methods such as adaptive parameter tuning methods~\cite{Gao2014optimal} have been developed in the literature.
We have provided a comprehensive review of the used methods to control the parameters of AFSA in Section~\ref{rev:param}.


\item \textit{Procedure and sub-functions}:
The main AFSA procedure for updating the positions of AFs is constructed by several so-called \emph{behaviors}.
Improvements on these behaviors or developing  new ones, is among the most important modifications of the AFSA.
These modifications of behaviors are reviewed in Section~\ref{rev:beh}.


\item \textit{Hybridization:}
Many works have hybridized AFSA with other methods to benefit from their strength and address their possible shortcomings.
This category can be further classified into: hybridization with population-based and non-population-based methods (e.g., clustering approaches).
Furthermore, the population-based methods can be further divided into merged and co-operative models.
On the one hand, in the merged models, the AFSA and the associated population-based methods are combined into a unified optimizer that uses the update rules and/or procedures of the combined algorithms.
On the other hand,  in the co-operative models, the algorithms work independently and share some information with each other.
For example, in some cooperative methods, the algorithms run in a serial manner and the output of one of them is used as the input of another~\cite{fang2017estimation}.
The hybrid AFSA methods are reviewed in Section~\ref{rev:hybrid}.

\item \textit{Multi-objective AFSA}:
To tackle multi-objective optimization, an optimizer needs to find a set of non-dominated solutions.
The AFSA is originally designed to perform global optimization in single-objective optimization problems.
Consequently, in order to tackle  multi-objective optimization problems, several works have modified the AFSA to adapt it to the requirements of such problems.
Section~\ref{rev:multi} gives a review of multi-objective AFSA models.

\item \textit{Dynamic AFSA}:
To solve optimization problems in dynamic environments, it is important that the optimizer can efficiently find the optimal solution and also track it over time after environmental changes.
Due to the specific challenges of dynamic optimization problems, optimizers that have been originally designed for static environments often fail to efficiently tackle such problems~\cite{Blackwell2006Multiswarms}.
Some works have modified the AFSA for tackling dynamic optimization problems
A detailed review of dynamic AFSA models are presented in Section~\ref{rev:dynamic}.

\end{itemize}

In the following subsections, we provide a detailed review of each modification, and Table~\ref{Table:summary} provides a summary of these modifications. 

\subsection{Parameters modification}
\label{rev:param}
{\color{black}In general, population-based algorithms are able to maintain a good trade-off between exploration and exploitation. Exploration refers to the ability to discover various solutions in the entire search space, while exploitation is the ability to search within the vicinity of the best solution found so far. These two abilities can be controlled through the algorithm parameters~\cite{crepinsek2011analysis}.
In the AFSA, the balance between exploration and exploitation is controlled by $\delta$, $v$, and $s$.
In particular, $v$ and $s$ affect the AFSA convergence performance, as both parameters are related to almost all behaviors.
Large $\delta$, $v$, and $s$ values increase the exploration ability, and reduce the stability and exploration ability in subsequent iterations. These settings enable AFs to change their locations with a large range. As such, they move faster toward the global optimum in the early stage and they are not able to search for more accurate positions during the subsequent stages, causing the AFs to oscillate around the global optimum until the termination criterion is met.
In contrast, while small $v$ and $s$ settings enable the AFs to improve local search capability, they can be trapped in local optimum solutions.
When $v$ is large, the AFs tend to execute the following and swarming behaviors more frequently than the preying behavior, and vice versa.\par

To alleviate these issues, adaptive $s$ and $v$ settings are proposed; which use large values at  early stages, and small values at later stages.}
For example, Yan et al.~\cite{yan2020application} used large $v$ and $s$ values for the preying behavior to obtain a better global search capability in the early stages, and consequently small $v$ and $s$ values for the following and swarming behaviors to improve the local search capability.
Gao et al.~\cite{gao2020twin} employed a broader $v$ setting for the foraging behavior to increase the global search capability and convergence speed.
Qin and Xu~\cite{qin2018adaptive} reduced $s$ and $v$ if no improvement is observed after certain number of iterations.
In~\cite{xi2019adaptive}, a predefined threshold was used to reduce $v$ at each iteration.

Besides, several studies used linear~\cite{Fei2014Motion,liu2016complexity} or non-linear~\cite{chen2016improved,el2015image,peng2018modification,tan2019normative,zhang2017fish} functions to dynamically adjust the AFSA parameters.
These functions can be based on the number of iterations, fitness value, distance or other conditions.
For example, Fei et al.~\cite{Fei2014Motion} used a linear function based on the current iteration number ($t$) to gradually reduce the $s$ and $v$ values as follows:
\begin{align}
\label{eq:vislin}
v^{(t)}=v_0-k_{v}.t,
\end{align}
\begin{align}
\label{eq:steplin}
s^{(t)}=s_0-k_{s}.t,
\end{align}
where $k_v$ and $k_s$ are the changing slope rates of $v$ and $s$, respectively.
Using (\ref{eq:vislin}) and (\ref{eq:steplin}) result in having larger $v$ and $s$ values in the early stages and smaller $v$ and $s$ values in the later stages, thus attempting to balance the trade-off between exploration and exploitation.

Similar to the linear functions, several studies~\cite{cheng2017parameter,feng2020analysis,Gao2014optimal,peng2018modification,zhang2017fish,zhang2016robot} adopted exponential functions based on the current ($t$) and maximum iteration ($t_\mathrm{max}$) numbers to gradually reduce $s$ and $v$, i.e.,
\begin{align}
\label{eq:visexp}
v^{(t)}=v_\mathrm{in}\times\alpha_v+v_\mathrm{min},
\end{align}
\begin{align}
\label{eq:stepexp}
s^{(t)}=s_\mathrm{in}\times\alpha_s+s_\mathrm{min},
\end{align}
where $v_\mathrm{in}$ and $s_\mathrm{in}$ are input values for calculating $v$ and $s$ over time, respectively, $v_\mathrm{min}$ and $s_\mathrm{min}$ are values computed according to the accuracy of the problem; and they indicate the minimum values that $v^{(t)}$ and $s^{(t)}$ can take, respectively, while $\alpha_v$ and $\alpha_s$ are exponential functions representing the changing parameter of the model operation.
As an example, Gao et al.~\cite{Gao2014optimal} used the following function:
\begin{align}
\label{eq:alpha}
\alpha_v=\alpha_s = \exp(-3 \cdot (\frac{t}{t_\mathrm{max}})).
\end{align}

Liu et al.~\cite{liu2019modified} adopted a function according to the problem, i.e., hydraulic-connectivity, to update $\alpha_v$ and $\alpha_s$ at each iteration, while Zhang et al.~\cite{zhang2016robot} used the attenuation function to update $v$ in (\ref{eq:visexp}) and the Gaussian distribution function to update $s$ in (\ref{eq:stepexp}), as follows:
\begin{align}
\label{eq:alphaa}
\alpha_v = \exp(-25 \cdot (\frac{t}{t_{max}})^k),
\end{align}
\begin{align}
\label{eq:alpha2}
\alpha_s = \frac{1}{\sigma \sqrt{2\pi}}e^{-(\frac{t}{t_{max}}-\mu)^2/2\sigma^2},
\end{align}
where $k$ is a value between 1 and 20, $\sigma$ represents the standard deviation, and $\mu=0$. {\color{black}Using Eqs.~(\ref{eq:alphaa}) and (\ref{eq:alpha2}) can minimize the values of $v$ and $s$ while allowing the AFs to move closely to the optimal solution during the iteration process.}\par

Mao et al.~\cite{mao2018comprehensive} introduced adaptive $v$ and $s$ based on five neighborhood AFs, as follows:
\begin{align}
\label{eq:visdis}
v_{i}^{(t)}=\mathcal{U}(0,1)\cdot \frac{1}{5}\cdot \alpha^{(t)} \cdot \sum_{j=1}^5d_{i,j}+v_\mathrm{min},
\end{align}
\begin{align}
\label{eq:stepdis}
s_{i}^{(t)}=\frac{1}{8}\cdot \alpha^{(t)}\cdot v^{(t)}_{i}+s_\mathrm{min},
\end{align}
where $v_\mathrm{min}$ and $s_\mathrm{min}$ are the minimum visual and step sizes, respectively, and $\alpha^{(t)}$ is defined as follows:
\begin{align}
\label{eq:alphadis}
\alpha^{(t)}=\exp(-r\cdot (\frac{t}{t_\mathrm{max}})^2),
\end{align}
where $r$ indicates the limited factor.\par

 Huang et al.~\cite{huang2021layout} used three adaptive $s$ to define the AFSA behaviors, as follows:
\begin{align}\label{at1}
    s_1^{(t)}=(1-\frac{t-1}{\beta_1-t_{max}})\cdot s^{(t-1)}, 
\end{align}
\begin{align}\label{at2}
    s_2^{(t)}=\beta_2\cdot s^{(t-1)}, 
\end{align}
and 
\begin{align}\label{at3}
    s_3^{(t)}=s^{(t-1)}+\frac{s_0}{\beta_3 \cdot t_{max}}, 
\end{align}
where $\beta_1$, $\beta_2$ and $\beta_3$ are the attenuation factors of the three adaptive steps.
{\color{black} Eqs. (\ref{at1}), (\ref{at2}) and (\ref{at3}) gradually decrease $s$, in order to increase the convergence speed and enhance the optimization precision. 
}

A multi-strategy AFSA (MSAFSA)~\cite{zhuang2019mechanical} model introduced an adaptive $s$, as follows:
\begin{align}
\label{eq:steplinn}
s^{(t)}=s_\mathrm{max}-\frac{t}{t_\mathrm{max}}(s_\mathrm{max}-s_\mathrm{min}),
\end{align}
where $s_\mathrm{max}$ and $s_\mathrm{min}$ are the maximum and minimum step sizes, respectively.\par

Yuan et al.~\cite{yuan2020asafsa} used the distance between the location of the $i$-th  AF and the best found position ($\vec{\mathbf{g}}$) to adjust $s_i$ in each iteration, as follows:
\begin{align}
\label{eq:dis}
s_i^{(t)} = \|\vec{\mathbf{g}}^{(t-1)}-\vec{x}_i\|\cdot\mathcal{U}(-1,1),
\end{align}
where $\vec{\mathbf{g}}^{(t-1)}$ is the best found position up to the last iteration.
{\color{black} This allows the AFs to move within a reasonable region, in a way to adjust the trade-off between the exploration and exploitation abilities.}\par

Similarly, a number of studies~\cite{li2018fault,lin2021kinematic,yan2020wireless} adjust the AFSA parameters adaptively according to the fitness value. In studies~\cite{lin2021kinematic,yan2020wireless} the following function to calculate $s$ based on the fitness value of the newly generated position and the current AF is proposed:
\begin{align}
\label{eq:stepad}
s_i^{(t)}=\left|1-\frac{f(\vec{x}_i)}{f(\vec{x}_j)}\right|\cdot s_i^{(t-1)}.
\end{align}

Yan et al.~\cite{yan2020wireless} also provided a function to calculate $v$ in each iteration for the $i$-th AF with respect to the distance of other AFs locating within the visual distance of the current AF, as follows:

\begin{align}
v_i^{(t)}=\frac{1}{\mathbb{N}(\vec{x}_i)}\sum_{j\in N_i} \|\vec{x}_i-\vec{x}_j\|,
\end{align}
where $\mathbb{N}(\vec{x}_i)$ is a set of neighbors of the $i$th AF (i.e., other AFs which lie inside the field of view ($v$) of the $i$-th AF), $f(\vec{x}_i)$ and $f(\vec{x}_j)$ are fitness value of current and newly generated AF, in which $f(\vec{x}_j)$ is better than $f(\vec{x}_i)$.
Since the value of $f(\vec{x}_i)/f(\vec{x}_j)$ is small in the early stages, $s_i$ is larger.
The value of $f(\vec{x}_i)/f(\vec{x}_j)$ increases with approaching to the optimal solution, resulting in a small value of $s_i$.
Meanwhile, Li et al.~\cite{li2018fault} introduced a $s$ function based on the fitness value of the current and previous iterations of the corresponding AF, and a conditional $v$ value.
If the improvement is higher than a user-defined threshold, a large $v$ setting is used, and vice versa.\par

{\color{black} In~\cite{gao2015triaxial,gao2015novel}, inspired by the random inertia weight of PSO~\cite{shi1998modified}, the $\alpha$ value is determined randomly in each iteration for the $i$-th AF:
\begin{align}\label{eq:inerr}
\alpha_i^{(t)}=\alpha_\mathrm{min}+\mathcal{U}(0,1)\cdot(\alpha_\mathrm{max}-\alpha_\mathrm{min}),
\end{align}
\begin{align}
v_i^{(t)}=v_i^{(t-1)}\cdot \alpha_i^{(t)},
\end{align}
where $\alpha_\mathrm{max}$ and $\alpha_\mathrm{min}$ indicate the maximum and minimum values of $\alpha$, respectively. 
Eq. (\ref{eq:inerr}) generates random values between $\alpha_\mathrm{max}$ and $\alpha_\mathrm{min}$, which can avoid local extreme points during the AF optimization process. 
}

Zhu et al.~\cite{zhu2020random} exploited the Lorentzian and exponential functions to adaptively adjust the values of $v$ and $s$, respectively, as follows:
\begin{align}
\vec{v}_i^{(t+1)}=\vec{v}_i^{(t)} \cdot (2\cdot (\frac{t}{t_\mathrm{max}})^2+1)^{-1},
\end{align}
\begin{align}
s_{i}^{(t+1)}=s_i^{(t)} \cdot e^{-\pi(\frac{t}{t_\mathrm{max}})^2}.
\end{align}

Liu et al.~\cite{liu2019improved} used the best found position $\vec{\mathbf{g}}$ to update $v$:
\begin{align}
v_i^{(t+1)}= v_i^{(t)}+ \mathcal{U}(-1,1)  \cdot\frac{\vec{\mathbf{g}}-\vec{x}_i^{(t)}}{\|\vec{\mathbf{g}}-\vec{x}_i^{(t)}\|}.
\end{align}


Kang et al.~\cite{kang2019optimization} divided the population into two groups, i.e., elite and normal, and assigned different adaptive $s$ and $v$ settings to each group.
{\color{black}Inspired by the ocean current power (OCP), the AFSAOCP~\cite{wang2016afsaocp} model divided the AFs into three groups, and assigned a different $s$ setting to each group.
This strategy enhances species diversity and allows the algorithm to escape from local optima during the iteration process.  }\par 


Moreover, a rational value of $\delta$ helps the AFSA to avoid the local optimal solutions~\cite{cai2010artificial}.
A large $\delta$ setting permits a lower crowding level, which causes the AFSA to jump out of local optima, but it reduces the convergence speed owing to executing the random behavior to avoid overcrowding.
In the subsequent stage, if the food location is too crowded, the AFs cannot swim to the area to probe for more accurate optimum solutions.
The experiment results in~\cite{Fernandes2009Fish} indicated that the convergence speed can be accelerated by gradually reducing $\delta$.
In this regard, Cheng and Xiang~\cite{cheng2017parameter} adaptively reduced the $\delta$ value as the number of iteration increased using:
\begin{align}
\label{eq:delta}
\delta^{t+1}=e^{(-\mu .t)} \delta _\mathrm{max},
\end{align}
where $\mu\in(0,1)$.\par

\subsection{Procedure and Sub-Functions}
\label{rev:beh}
{\color{black} This category of modifications aims to improve the AFSA performance by modifying its behaviors, such as incorporating the updating mechanisms from other population based methods into the AFSA behaviors or introducing new ones. }
Apart from the three behaviours of the original AFSA, \textit{leaping}~\cite{farzi2009efficient} and \textit{swallow}~\cite{cheng2009novel} behaviours have been introduced to improve the performance of the algorithm.
Leaping behaviour randomly selects a fraction of the AFs, and re-initialize them within their visual distance.
Note that the best AF should not be included among the selected AFs.
The leaping behavior is executed if after a certain number of iterations, the objective function of the best AF is not improved or the improvement is smaller than a threshold~\cite{farzi2009efficient}.
The MSAFSA~\cite{zhuang2019mechanical} model introduced both the swallow~\cite{cheng2009novel} and leaping behaviors to reduce complexity and escape from the local optima, respectively.
It applies the swallow behavior after a certain number of iterations, i.e., when the fitness value of an AF is smaller (for maximum problem) or larger (for minimum problem) than a user-defined threshold, it is swallowed.
In addition, it activates the leaping behavior in situations when no improvement is observed in the objective function of two adjacent iterations.
When the fitness value of an AF is $\beta$ times more than the global optimum, the AF is aborted to release the memory and reduce the model complexity.\par

{\color{black}Since the convergence speed in the later stage of AFSA is too slow, i.e., increasingly more AFs perform invalid search that requires more  time . To alleviate this issue, the population inhibition behavior was proposed in~\cite{huang2015log}. In this behavior, after a certain number of iterations, big fishes eat small fishes, and the occupied space of small fishes is cleared, helping to accelerate the convergence speed.  
Li et al.~\cite{li2015quantum} modified the behaviors with a Bloch spherical-based quantum coding to realize individual evolution quantity.  With a three-chain quantum bit encoding mechanism, each AF simultaneously searches three new locations during each iteration.  Each search behavior must involve measurement and projection based on a bit rotation matrix structure. Although this mechanism significantly increases the computational load, it can produce better results as compared with the original AFSA.}
Jia et al.~\cite{jia2020privacy} used a bulletin board to show the distance between the AFs and each privacy-sensitive service selection in an internet-of-things (IOT) environment.\par

{\color{black}Gao et al.~\cite{gao2020twin} developed an \textit{elimination} and \textit{regeneration} mechanism to improve the quality of AFs. As such, AFs whose fitness values are lower than a user-defined threshold are removed, and the same number of AFs are generated for replacement purposes. This strategy helps the AFs to conduct effective search and avoid being trapped in local optima.}
In addition, a Cauchy mutation is applied to the foraging behavior of the AFSA to jump out of the local optima. 
If the AF cannot move toward a better position after $\mathfrak{t}$ attempts of executing the foraging behavior, instead of performing random behavior, the Cauchy mutation is used to mutate the $i$th AF as follows:
\begin{align}
\label{eq:viscauch}
 v_i^{(t)}=v_0\cdot(1+\frac{t}{\mathfrak{t}}),  
\end{align}
\begin{align}
\label{eq:poscauch}
 \vec{x}_{p,i}=\vec{x}_i^{(t)}(1+c),
\end{align}
where $c$ is a variable that obeys the Cauchy distribution. 
The use of~(\ref{eq:viscauch}) and~(\ref{eq:poscauch}) gradually expands the search space, thus enabling the AF to escape from the local optimum. 
In~\cite{dawei2015wireless}, {\color{black} the Gaussian distribution function was used to initialize the AFs with the preying behaviour, in order to obtain a more precise location in the feeding stage. The preying behavior can be considered as a random movement that depends on the current location of the AF and probability of moving to the next position.}
On one hand, the preying behavior can be considered as a random movement that depends on the current location of the AF and probability of moving to the next position.
In this regard, Peng et al.~\cite{peng2018modification} applied a l\' {e}vy search strategy~\cite{pavlyukevich2007levy}, which is an effective strategy with random movement, to the preying behavior, i.e.,
\begin{align}
\label{eq:levy}
 \vec{x}_{p,i}=\vec{x}_i^{(t)}+\gamma \cdot L(\lambda),
\end{align}
where $\gamma$ is the step factor between 0 and 1, $L(\lambda)$ is a random vector generated by the l\' {e}vy strategy, as follows:
\begin{align}
\label{eq:levyy}
L(\lambda)=\frac{\phi\cdot \mu}{|r|^{1/\beta}}\cdot \left(\vec{x}_i^{(t)}-\vec{\mathbf{g}}\right), 
\end{align}
where $1<\lambda<3$, $0<\beta<2$, $\vec{\mathbf{g}}$ is the best found position, $\mu=t^{-\lambda}$,  $\mu \sim \mathcal{N}(0,\phi^2)$ and $r \sim \mathcal{N}(0,1)$ obey the normal distribution, and:
\begin{align}
 \phi=\bigg\{ \frac{\Gamma(1+\beta)\sin (\pi \beta/2)}{\Gamma[(1+\beta)/2]\beta.2^{(\beta-1)/2}}\bigg\}^{1/\beta},
\end{align}
where $\Gamma$ indicates the standard Gamma function.
In addition, the moving strategy of the AF is integrated into the preying and following behaviors, as follows:
\begin{align}
\vec{x}_{p,i}=w^{(t)}\cdot \vec{x}_{i}^+\beta_{ij}(\vec{x}_j,\vec{x}_i^{(t)})+\alpha \cdot (\mathcal{U}(-1,1)-0.5),
\end{align}
where $\alpha$ is a step factor, $\alpha \cdot (\mathcal{U}(-1,1)-0.5)$ prevents falling into the local optimum, and $\beta_{ij}(\vec{x}_j,\vec{x}_i^{(t)})$ and $w^{(t)}$, respectively, are the attraction level and inertia weight, defined as follows:
\begin{align}
\beta_{ij}(\vec{x}_j,\vec{x}_i^{(t)})=\beta_0\cdot e^{-\gamma r_{ij}},
\end{align}
\begin{align}
\label{eq:inertia}
w^{(t)}=w_\mathrm{max}-\frac{w_\mathrm{max}-w_\mathrm{min}}{t_\mathrm{max}}\cdot t,
\end{align}
where $r_{ij}$ is the the Cartesian distance between the $i$-th and the $j$-th AFs, $\beta_0$ is the attraction degree of an AF at $r_{ij}=0$, $w_{max}$ and $w_{min}$ are the maximum and minimum inertia weights. 
Note that for the preying behavior, $\vec{x}_j$ is obtained by~(\ref{eq:prey1}), while the following behavior serves as a better companion within the visual distance of the current AF. 
Using the l\' {e}vy search and the FA moving strategies help the model to avoid local optima. 

Zhu et al.~\cite{zhu2020random} divided the population into several sub-populations.
Each sub-population searches independently within its search area, and the best individuals of all sub-populations are allowed to interact efficiently.
As such, new positions are generated based on two individuals as follows:
\begin{align}
\begin{cases}
\vec{x}_i^{'(t)}=\frac{1}{2}((1+\mathcal{U}(0,1)) \vec{x}_i^{(t)}+(1-\mathcal{U}(0,1))\vec{x}_j^{(t)}))\\
\vec{x}_j^{'(t)}=\frac{1}{2}((1-\mathcal{U}(0,1)) \vec{x}_i^{(t)}+(1+\mathcal{U}(0,1))\vec{x}_j^{(t)}))
\end{cases},
\end{align}
where $\vec{x}_i^{(t)}$ and $\vec{x}_j^{(t)}$ are two different AFs. Then, replacement take place under the following conditions:
\begin{align}
\begin{dcases}
\vec{x}_i^{(t+1)}=\vec{x}_i^{'(t)},& \text{if}~f(\vec{x}_i^{'(t)})<f(\vec{x}_i^{(t)})\\
\vec{x}_j^{(t+1)}=\vec{x}_j^{'(t)},& \text{if}~f(\vec{x}_j^{'(t)})<f(\vec{x}_j^{(t)})
\end{dcases}.
\end{align}


The original AFSA uses a bulletin to record the best AF.
This is because the random behavior can move the best solution to a sub-optimum location.
As such, other AFs cannot use the benifits of the current best solution to move toward the best position; consequently, they do not effectively search within the vicinity of the local position.
To overcome this issue and improve the local search ability of the AFSA, several studies implement the global best or local best position into the behaviors of the AFSA.
For example, 
Fei et al.~\cite{Fei2014Motion} modified the preying behavior by combing the current AF and the best found AF (i.e., $\vec{\mathbf{g}}$), as follows:
\begin{align}
\vec{x}_{p,i}=\vec{x}_i^{(t)}+s_i^{(t)} \cdot \mathcal{U}(0,1)\cdot \left(\frac{\vec{x}_j-\vec{x}_i^{(t)}}{\|\vec{x}_j-\vec{x}_i^{(t)}\|}+\frac{\vec{\mathbf{g}}-\vec{x}_i^{(t)}}{\|\vec{\mathbf{g}}-\vec{x}_i^{(t)}\|}\right).
\end{align}

The AFSA model in~\cite{gao2015triaxial,gao2015novel} consisted of two behaviors, i.e., individual and group behaviors.
The individual behavior is based on the preying behavior.
Specifically, it generates a new position $\vec{x}_j$ by the praying behavior. 
If $\vec{x}_j$ is better than $\vec{x}_i^{(t)}$, the AF moves toward $\vec{x}_j$.
Otherwise, it randomly generates a new position.
The group behavior generates a new AF based on the center position if the fitness value of the center position is better than that of the current AF, otherwise, it generates a new AF based on the best found position, i.e.,:
\begin{align}
\vec{x}_i^{(t+1)}=
\begin{dcases}
\vec{x}_i^{(t)}+v_i^{(t)}\cdot \mathcal{U}(0,1) \cdot \frac{\vec{x}_c-\vec{x}_i^{(t)}}{\|\vec{x}_c-\vec{x}_i^{(t)}\|}, & \text{if}~f(\vec{x}_c)>f(\vec{x}_i^{(t)})\\
\vec{x}_i^{(t)}+v_i^{(t)}\cdot \mathcal{U}(0,1) \cdot \frac{\vec{\mathbf{g}}-\vec{x}_i^{(t)}}{\|\vec{\mathbf{g}}-\vec{x}_i^{(t)}\|}, & \text{Otherwise},
\end{dcases}.
\end{align}

Inspired by the PSO~\cite{shi1998modified}, Huang and Chen~\cite{huang2015log} introduced an inertia weight-based adaptive movement behavior and a log-linear scheme to select the behaviors, i.e., 
\begin{align}
\label{eq:iner}
w^{(t)}=0.6\cdot e^{(-10\cdot(1-P_r^{(t)}))},
\end{align}
where $P_r^{(t)}$ is the log-linear function derived from three feature functions i.e., diversity function, dimensional distribution function, and average distance metrics function. 
Then, they formulated new preying~(\ref{eq:preyy}), following~(\ref{eq:swarmm}) and swarming~(\ref{eq:followw}) behaviors according to~(\ref{eq:iner}), as follows:   
\begin{align}
\label{eq:preyy}
\vec{x}_{p,i}=(1-w^{(t)})\cdot \vec{x}_i^{(t)}+w^{(t)} \cdot s_i^{(t)} \cdot \mathcal{U}(0,1) \cdot \frac{\vec{x}_j^{(t)}-\vec{x}_i^{(t)}}{\|\vec{x}_j^{(t)}-\vec{x}_i^{(t)}\|},
\end{align}
\begin{align}
\label{eq:swarmm}
\vec{x}_{f,i}=(1-w^{(t)})\cdot \vec{x}_i^{(t)}+w^{(t)}\cdot s_i^{(t)}\cdot \frac{\vec{x}_c-\vec{x}_i^{(t)}}{\|\vec{x}_c-\vec{x}_i^{(t)}\|}, 
\end{align}
\begin{align}
\label{eq:followw}
\vec{x}_{s,i}=(1-w^{(t)})\cdot \vec{x}_i^{(t)}+w^{(t)}\cdot s_i^{(t)}\cdot \frac{\vec{\mathbf{g}}-\vec{x}_i^{(t)}}{\|\vec{\mathbf{g}}-\vec{x}_i^{(t)}\|}. 
\end{align}

Zhang et al.~\cite{zhang2016robot} used the inertia weight factor similar to~(\ref{eq:inertia}) to improve the movement of AFs in the search space, and balance the trade-off between exploration and exploitation.
Then, they adopted~(\ref{eq:swarmm}) and~(\ref{eq:followw}) to perform swarming and following behaviors, respectively.
In addition, the preying behavior is formulated as follows: 
\begin{align}
\label{eq:preyyy}
\vec{x}_{p,i}=(1-w^{(t)})\cdot \vec{x}_i^{(t)}+w^{(t)}\cdot s_i^{(t)}\cdot \mathcal{U}(-1,1).
\end{align}

Cheng and Lu~\cite{cheng2018research} modified the preying behavior of the AFSA based on the velocity-displacement strategy of the PSO.
As such, the velocity of each AF is computed, as follows:
\begin{align}
\label{eq:velo}
\vec{\mathsf{v}}_i^{(t+1)}=w\cdot\vec{\mathsf{v}}_i^{(t)}+c_1\mathcal{U}(0,1)(\vec{\mathbf{p}}_i^{(t)}-\vec{x}_i^{(t)})+c_2\mathcal{U}(0,1)(\vec{\mathbf{g}}-\vec{x}_i^{(t)}),
\end{align}
where $c_1$ and $c_2$ are the learning rates, $w$ is the inertia wight, and $\vec{\mathbf{p}}_i^{(t)}$ is the best position of the $i$-th AF.
Then, a new position is generated, i.e.:
\begin{align}
\label{eq:upd}
\vec{x}_{p,i}=\vec{x}_i^{(t)}+\alpha \cdot \vec{\mathsf{v}}_i^{(t+1)},
\end{align}
where $\alpha$ is a constraint factor.

In a similar way, Cao et al.~\cite{cao2017modified} modified the swarming and following behaviors of the AFSA. 
Specifically, for the swarming behavior, they replaced $\vec{x}_c$ with $\vec{\mathbf{p}}_i^{(t)}$ in~(\ref{eq:velo}). 
For the following behavior, they replaced the best AF in the visual range with $\vec{\mathbf{p}}_i^{(t)}$ in~(\ref{eq:velo}), leading to a new position, i.e.,
\begin{align}
\vec{x}_{f,i}=\vec{x}_i^{(t)}+s_i^{(t)} \cdot \frac{\vec{\mathsf{v}}_i^{(t)}}{\|\vec{\mathsf{v}}_i^{(t)}\|}.
\end{align}

Mao et al.~\cite{mao2018comprehensive} applied the velocity and best found position of the PSO with extended memory (PSOEM)~\cite{duan2011simulation} to the AFSA behaviors.
As such, the velocity of each AF pertaining to their swarming, memory and communication behaviors are updated according to (\ref{eq:sw}), (\ref{eq:mem}) and (\ref{eq:com}), respectively.
Then, the new positions are generated based on formulation in~(\ref{eq:upd}), i.e., 
\begin{align}
\label{eq:sw}
\vec{\mathsf{v}}_i^{(t+1)}=w\vec{\mathsf{v}}_i^{(t)}+s_i^{(t)} \cdot \mathcal{U}(0,1)\cdot\frac{\vec{x}_c-\vec{x}_i^{(t)}}{\|\vec{x}_c-\vec{x}_i^{(t)}\|}, 
\end{align}

\begin{align}
\label{eq:mem}
\vec{\mathsf{v}}_i^{(t+1)}=w\vec{\mathsf{v}}_i^{(t)}+s_i^{(t)}\cdot \mathcal{U}(0,1)\cdot\frac{[\xi^{(t)} (\vec{\mathbf{p}}_i^{(t)}-\vec{x}_i^{(t)})+\xi^{(t-1)} (\vec{\mathbf{p}}_i^{(t-1)}-\vec{x}_i^{(t-1)})]}{\|[\xi^{(t)} (\vec{\mathbf{p}}_i^{(t)}-\vec{x}_i^{(t)})+\xi^{(t-1)} (\vec{\mathbf{p}}_i^{(t-1)}-\vec{x}_i^{(t)-1})]\|},
\end{align}

\begin{align}
\label{eq:com}
\vec{\mathsf{v}}_i^{(t+1)}=w\vec{\mathsf{v}}_i^{(t)}+s_i^{(t)}\cdot \mathcal{U}(0,1)\cdot\frac{[\xi^{(t)} (\vec{\mathbf{g}}-\vec{x}_i^{(t)})+\xi^{(t-1)} (\vec{\mathbf{g}}^{(t-1)}-\vec{x}_i^{(t-1)})]}{\|[\xi^{(t)} (\vec{\mathbf{g}}-\vec{x}_i^{(t)})+\xi^{(t-1)} (\vec{\mathbf{g}}^{(t-1)}-\vec{x}_i^{(t-1)})]\|},
\end{align}
where $\vec{\mathbf{p}}_i^{(t-1)}$ and $\vec{\mathbf{g}}^{(t-1)}$ are, respectively, the best position of the $i$-th AF and best found position after $(t-1)$ iteration, $\xi^{(t)}$ and $\xi^{(t-1)}$ are the effective factors of the extended memory at iterations $t$ and $(t-1)$, respectively.

\subsection{Hybrid AFSA}
\label{rev:hybrid}
{\color{black}Many hybrid models of the AFSA combined with other methods have been proposed.  The goal of hybrid models is to exploit the strengths of each method by integrating them into one single model.}
Hybrid AFSA-based models can be divided into two groups: \emph{(i)} hybrid models with population-based methods; and \emph{(ii)} hybrid models with non-population-based methods.
These two groups are reviewed as follows.

\subsubsection{Population-based methods}
\label{rev:hybrid:pop}
Hybrid AFSA models with population-based algorithms can be further divided into two groups.
In the first group, two or more methods are merged into a unified model, while in the second group, the methods co-operate in parallel or serial to find the global optimum.

\begin{figure}[tb!]
 \begin{center}
 \includegraphics[scale=0.850]{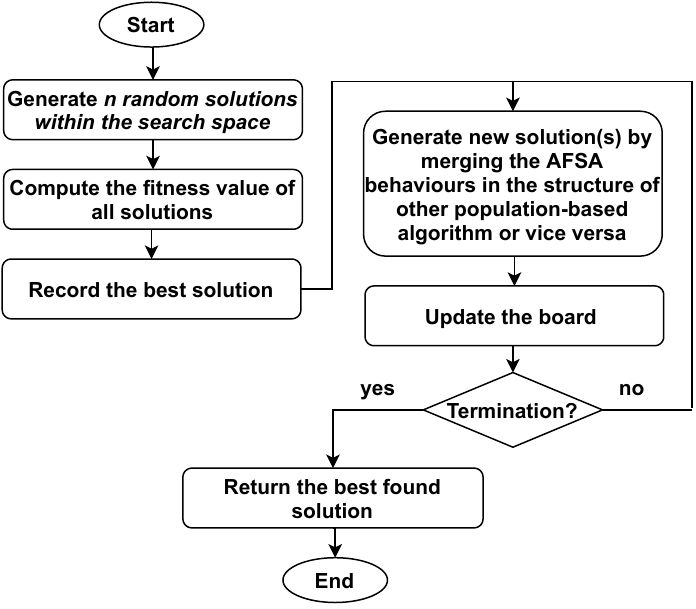}
  \end{center}
  \vspace*{-0.5cm}
  \caption{\color{black}Flowchart of the merged hybrid AFSA models.  The AFSA dynamic is embedded within the structure of other population-based models, or vice versa, to generate new solutions iteratively until the termination condition is satisfied.}
  \label{Fig:hybrid1}
\end{figure}

\textbf{Merged models:} {\color{black} These models, as shown in Fig.~\ref{Fig:hybrid1}, generate new solutions by merging the AFSA dynamic within the structure of other population-based algorithms, or vice versa.} 
PSO has been widely merged with the AFSA.
An AFSA-PSO model~\cite{wei2019optimization} to optimize the basic structure of the deep belief network was proposed.
The AFSA's behaviors are implemented in the PSO structure to improve the movement of particles.
Yassen et al.~\cite{yaseen2018optimization} integrated the follow and swarm behaviors of the AFSA in PSO.
To achieve this, PSO is used to find the global best and local best positions, and then, the particles are divided into two sub-populations.
The following behavior of the AFSA is applied to one group, and the swarm behavior is carried out on the other group.
Finally, the best particle obtained by these two behaviors is compared with the recorded particle in memory.
If the new particle is better than the recorded one, replacement takes place.
In MPSO~\cite{yan2020anovel}, the preying and random behaviors of the AFSA were embedded in PSO, in order to increase its local search capability for the purpose of K-means clustering optimization.
In~\cite{cheng2018research}, the velocity and position update of PSO was used to improve the preying behavior of the AFSA.
In addition, a non-uniform mutation operator was used as a reference to enhance the local searchability.
CIAFSA~\cite{mao2018comprehensive} combined PSO with an extended memory (PSOEM)~\cite{duan2011simulation} and the AFSA.
CIAFSA implemented the velocity of particles into the behaviors of the AFSA.
Later, normative AFSA (NFSA)~\cite{tan2019normative}, namely a hybrid model of PSOEM-FSA and normative knowledge~\cite{reynolds2004cultural}, was proposed.
Yuan and Yang~\cite{yuan2019study} implemented the swarming and following behaviors of the AFSA in the search strategy of PSO.
Zhang et al.~\cite{zhang2017adaptive} implemented the speed-up behavior of PSO in the AFSA.
MSAFSA~\cite{zhuang2019mechanical} implemented the search strategy of DE and PSO in the AFSA's behaviors.
Cao et al.~\cite{cao2017modified} applied the velocity-displacement strategy of PSO in the AFSA to increase the convergence speed.

The AFSA has also been merged with other population-based algorithms.
He et al.~\cite{he2019multi} implemented the crossover and mutation strategies of DE in the later stage of the AFSA to improve the global search ability by escaping from local optima. Hajisalem and Babaei~\cite{hajisalem2018hybrid} 
{\color{black} introduced ABC-AFSA for anomaly detection by incorporating the preying behavior of the AFSA into ABC. Since the preying behavior can determine the direction of food, it helps the model to avoid pre-matured convergence.}
Zhang et al.~\cite{zhang2021improved} proposed FDMABC for function optimization by exploiting the AFSA crowding factor to present the ABC algorithm from falling into local optima. 
SA-IAFSA~\cite{he2016hybrid}, namely  a hybrid model of the AFSA and simulated annealing (SA), was  proposed to device a fuzzy-based clustering approach. SA-IAFSA used the AFSA with the leaping behavior, and it applied the simulated annealing (SA) in the preying behavior of the AFSA to increase stability and convergence speed of the model.

{\color{black}AF-GBFO~\cite{zheng2020hybrid}  combined the following and swarming behaviors of the AFSA with the chemotaxis part of the bacterial foraging algorithm (BFO).  The bacteria positions are updated based on an evaluation of their own positions as well as others to solve the poor convergence speed of the BFO and avoid being trapped in local optima in dealing with some complex problems.}
While, AFSA-BFO~\cite{fei2017application} adapted the BFO in the later stage of the AFSA to improve its local search ability.
{\color{black}LFFSA~\cite{peng2018modification} applied the moving strategy of the FA in the preying and following behaviors of the AFSA to address the issue of random moving after determining the direction in AFSA.}
In addition, the l\' {e}vy flight strategy is used during search to improve the preying behavior and avoid trapping into local optimum.
In a similar way, Xian et al.~\cite{xian2018novel} integrated the chemotactic behavior of BFO in the preying behavior of the AFSA, and a mutation strategy-based l\' {e}vy flight search.

\begin{figure}[tb!]
 \begin{center}
 \includegraphics[scale=0.850]{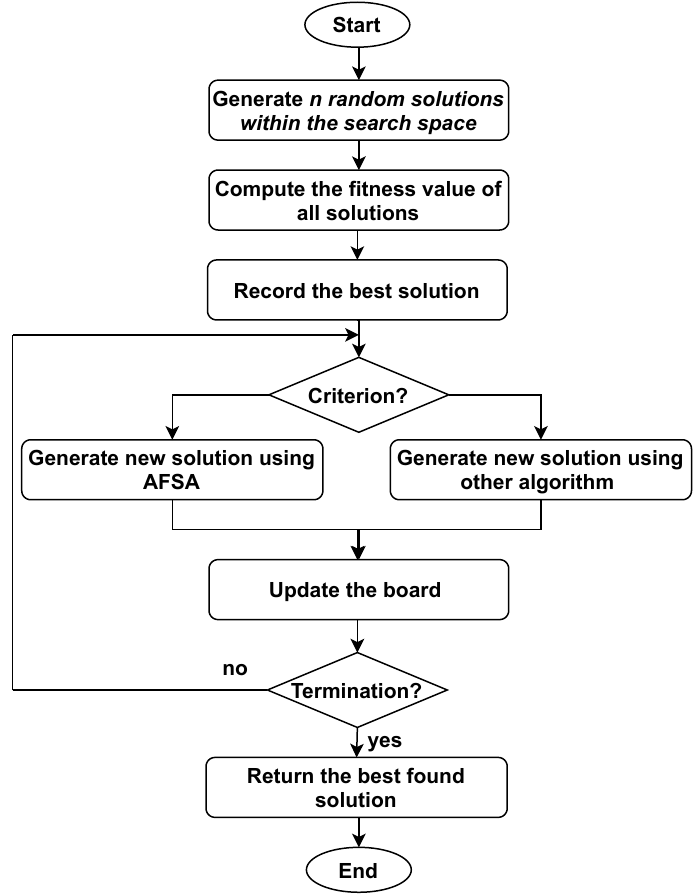}
  \end{center}
  \vspace*{-0.5cm}
  \caption{\color{black}Flowchart of the co-operative hybrid AFSA models. Each co-operative algorithm is utilized separately (either in parallel or in series) to find an optimal solution.}
  \label{Fig:hybrid2}
\end{figure}

\textbf{Co-operative models:} {\color{black} These models (as shown in Fig.~\ref{Fig:hybrid2}) employ the AFSA and other population based methods separately to generate new solutions, and combine the generated solution based on a formulated criterion.} 
A hybrid model of the AFSA with PSO based on Gaussian learning to improve the parameters of active disturbance rejection was proposed by Kang et al.~\cite{kang2019optimization}.
This method divides the solutions into several groups and then ranks the members of each group based on their fitness value.
The top 20\% of individuals are defined as the elites, and the rest as normal.
In addition, Gaussian learning is applied to further improve the position of the final particles. Fang et al.~\cite{fang2017estimation} proposed a hybrid model of the AFSA with PSO to detect the onset of ultrasonic signals.
This model, which uses a multi-modal objective function, firstly employs the AFSA to find all possible search space.
After extracting the optimal solutions in every space, PSO is used for exploitation.
{\color{black}
In contrast, HAFPSO~\cite{kanimozhi2021hybrid} first applies PSO to allow the AFSA to utilize the final best population from PSO as its initial population for performing the local search. Other hybrid models of PSO and the AFSA~\cite{jiang2017application,zhang2021parameter} use both algorithms separately to search for the best position, i.e., the global best and local best in PSO, and the best AF in the AFSA, before comparing the obtained best positions at the end of each iteration.  If a better position is found, the current best AF is replaced and the overall result of the AFSA is updated.
}

RCGA-AFSA~\cite{fang2014hybrid}, which is a hybrid model of real-coded genetic algorithm (RCGA) and AFSA, for solving short-term hydrothermal scheduling is proposed.
RCGA-AFSA uses RCGA as a global search algorithm to explore better solution spaces, and then AFSA is applied as a local search to exploit an accurate optimal solution.
Guo et al.~\cite{guo2016application} propose a hybrid model of AFSA, ACA, and backpropagation network (BPN), i.e., AFSAACA-BPN, to control thermal error.
It, firstly, uses AFSA to adjust BPN's weight, and then, AFs with top 5\% concentration is used to initialize the pheromone value of ACA.
Li et al.~\cite{li2016hybrid} develop a hybrid parallel model of AFSA and ABC.
This model, in the early stage, randomly divides swarms into two groups, and then applies different search strategy, i.e., AFS and ABC, to each group.
In the late stage, DN-AFS~\cite{li2013artificial}  and RP-ABC~\cite{binghui2006random}, which are variants of AFSA and ABC, are used to further improve the optimization performance.
In~\cite{cheng2018research}, the Metropolis rule of the simulated annealing (SA) is implemented in AFSA to avoid model from premature convergence.
In~\cite{li2018fault}, chaos adaptive AFSA for fault detection optimization problems is proposed.
This model applies Chaos search at the end of each iteration to further improve the best AF that is stored in the bulletin board.

{\color{black}In~\cite{fei2021location}, a hybrid model of AFSA and bacterial colony Chemotaxis (BCC), namely BCC-AFSA, for selection and optimization of distribution center location was proposed. 
BCC-AFSA applies BCC when the results of the AFSA do not change significantly after a certain number of iterations, in order to improve its global optimization. On the other hand, RNA-AFSA~\cite{zhang2021research} uses the characteristics of RNA (ribonucleic acid) to solve the oscillation problem in the late stage of the AFSA algorithm. In~\cite{upadhyay2021periodic}, a hybrid model of the AFSA and flower pollination algorithm (FPA) for obtaining the best dense cluster was introduced. This model employs either the AFSA or FPA at each iteration to find the best solution based on a defined criterion.  
}

\subsubsection{Non-population-based methods}
\label{rev:hybrid:non}
The AFSA has also been integrated with non-population-based methods.
In~\cite{xi2019adaptive,ma2015hybrid}, hybrid models of the AFSA with fuzzy \emph{C}-means (FCM) for clustering were proposed.
The AFSA is used to avoid FCM from being trapped in local optima.
Serapi\~{a}o et al.~\cite{serapiao2016combining} combined a variant of the AFSA proposed in~\cite{bastos2008novel} with \emph{K}-means and \emph{K}-harmonic methods for data clustering.
El-Said~\cite{el2015image} combined the AFSA with FCM for image quantization.
GAFSA-SVR~\cite{liu2016study}, namely a hybrid model of GAFSA~\cite{wang2015blind} and the support vector regression (SVR), for prediction of network traffic was proposed.
This model uses GAFSA to optimize the parameters of SVR.
Hybrid fish-swarm logic regression (FSLR) was proposed in~\cite{zhang2013identifying}.
{\color{black} 
On the other hand, Krishnaraj et al.~\cite{Krishnaraj20212artificial} proposed a clustering technique based on the AFSA and hill climbing for throughput maximization in wireless multimedia sensor networks.\par

In~\cite{hua2021misalignment,xian2021early}, the AFSA was adopted to optimize the SVM parameters. }
In~\cite{sathya2017hybrid}, the AFSA was combined with an artificial neural network (ANN) to find the best weights.
{\color{black}
Goluguri et al.~\cite{goluguri2021rice} used the AFSA to optimize the weights of a deep convolutional neural network for identifying rice diseases. 
}
In~\cite{zhuang2019mechanical}, MSAFSA-SVR was introduced to calculate the stability of rocks in tunnel engineering. 
MAFSA-SVR~\cite{yan2020application} was also used to predict the heat transfer capacity of radiators.
{\color{black} In~\cite{he2021multi}, the AFSA was applied to train a wavelet neural network (WNN).}
In~\cite{jia2020aparametric}, the AFSA was used for feature selection and parameter optimization of a random forest for detecting Cervical cancer.
In~\cite{gao2020twin}, a hybrid model of twin SVM (TSVM) with an improved AFSA model for flame recognition problem was proposed, in which the improved AFSA model was used to solve the parameter selection problem of TSVM.

ZAFSA~\cite{wang2018estimation}, namely a hybrid model of the AFSA and normal distribution, was proposed to evaluate the interfacial heat transfer coefficient.
ZAFSA, firstly, diminishes the search space using the normal distribution before applying the AFSA to find the best solution.
A hybrid model of the AFSA with singular value decomposition (SVD) for enhancing weak digital signals was  proposed in~\cite{zhang2019adaptive}.
{\color{black}
Tirkolaee et al.~\cite{babaee2020fuzzy} formulated a hybrid model based on the AFSA and an interactive fuzzy solution technique for flowshop scheduling.  The AFSA is able to provide the Pareto optimal solutions. CPAFSA~\cite{zhou2021chaotic} was proposed to optimize water quality monitoring sensor networks by creating new underwater sensor coverage algorithms. Specifically, CPAFSA uses chaotic selection for parameter initialization and integrates the global search capabilities of parallel operators. In addition, CPAFSA applies the elite selection to avoid local optimization and solve the problem of 3D target coverage.  AFSA-CA~\cite{huang2021optimization} was developed to optimize parameters of an urban growth model.

}

\subsection{Multi-objective AFSA  }
\label{rev:multi}
MOOP considers problems involving more than one objective function to be optimized.
It aims to find a trade-off between two or more conflicting objectives.
Several studies have been adopted AFSA and its variants to solve MOOP.
For examples, Sun et al.~\cite{sun2017application} proposed a multi-objective optimization algorithm based on the AFSA for undertaking the trajectory problems.
Later, Xu et al. proposed an iterative deletion AFSA (IDAFSA) to solve multi-objective problems~\cite{xu2019integrated}.
IDAFSA integrated the global optimal solution into the behaviors of the original AFSA.
In addition, the Pareto optimally proposed by~\cite{zhang2017pareto} was  adopted as a sorting mechanism to select a Pareto optimal solution set.
Ma and He~\cite{ma2019green} combined an adaptive GA~\cite{kusakci2014adaptive} with the AFSA to solve green wave traffic control.
When the recorded solution in the AFSA remained unchanged or improved slightly, the adaptive crossover and mutation operators were used to retain the optimal solution state of the AFs and conduct mutation pertaining to a small number of dimensions for other AFs.
{\color{black}In addition, Liu at al.~\cite{liu2020solving} developed a Pareto AFSA embedded with the GA to solve the urban electric transit network problem. 

}

\subsection{Dynamic AFSA}
\label{rev:dynamic}
 The AFSA also has been used to tackle dynamic optimization problems (DOPs).
In DOPs, the search space changes over time, due to the changes in the objective function, constraints, variable interactions, and/or the number of variables~\cite{Mavrovouniotiz2017asurvey}.
A DOP requires an algorithm to not only find the global optimum, but also track it after environmental changes. 
In view of the specific challenges of DOPs, such as diversity loss and outdated memory, optimization algorithms that have been designed to tackle static optimization problems cannot be directly used for tackling DOPs.
The AFSA is no exception to this maxim.\par

\begin{table}[tb]
\centering
\caption{\label{Table:summary} {\color{black}A summary of the modifications AFSA. }}
   \begin{adjustbox}{width=01.1\textwidth}
    \begin{tabular}{l c c}
    \toprule
    Modification & & Study \\ 
      \midrule
    \multirow{2}{*} {Parameters} & & \cite{Gao2014optimal,zhuang2019mechanical,yan2020application,qin2018adaptive,xi2019adaptive,Fei2014Motion,liu2016complexity,peng2018modification,zhang2017fish,el2015image,chen2016improved,tan2019normative,zhang2016robot,feng2020analysis,cheng2017parameter,lin2021kinematic},  \\
    &&\cite{liu2019modified,mao2018comprehensive,yuan2020asafsa,yan2020wireless,li2018fault,gao2015triaxial,gao2015novel,zhu2020random,kang2019optimization,wang2016afsaocp,dawei2015wireless,li2016hybrid,ma2015hybrid,yazdani2011color,huang2021layout}   \\
    \midrule
    Procedure and & &\cite{Gao2014optimal,zhuang2019mechanical,fang2017estimation,qin2018adaptive,Fei2014Motion,peng2018modification,zhang2017fish,chen2016improved,zhang2016robot,gao2015triaxial,gao2015novel,zhu2020random,liu2019improved,huang2015log},  \\
    sub-functions & &\cite{li2015quantum,jia2020privacy,cheng2018research,cao2017modified,zhang2017adaptive,he2019multi,he2016hybrid,fei2017application,xian2018novel,serapiao2016combining,liu2016study,wang2015blind,yazdani2011color,li2021computation}  \\
    \midrule
   \multirow{4}{*} { Hybrid} &  Population & \cite{fang2017estimation,peng2018modification,chen2016improved,mao2018comprehensive,li2018fault,kang2019optimization,cheng2018research,cao2017modified,wei2019optimization,yaseen2018optimization,yuan2019study,zhang2021improved,upadhyay2021periodic,kanimozhi2021hybrid},\\
   &based&\cite{zhang2017adaptive,he2019multi,hajisalem2018hybrid,he2016hybrid,zheng2020hybrid,fei2017application,xian2018novel,jiang2017application,fang2014hybrid,guo2016application,li2016hybrid,fei2021location,zhang2021research}  \\
   \cmidrule{3-3}
     & Non-population & \cite{yazdani2013new,Gao2014optimal,yan2020application,xi2019adaptive,liu2016complexity,tan2019normative,gao2015novel,ma2015hybrid,serapiao2016combining,liu2016study,huang2021optimization,he2021multi,goluguri2021rice,Krishnaraj20212artificial}, \\
     &based&\cite{zhang2013identifying,sathya2017hybrid,wang2018estimation,zhang2019adaptive,yazdani2011color,yazdani2010new,yazdani2010CA,hua2021misalignment,xian2021early,zhou2021chaotic,babaee2020fuzzy}\\
     \midrule
     Multi-objective &&\cite{babaee2020fuzzy,liu2020solving,ma2019green,sun2017application,xu2019integrated} \\
     \midrule
     Dynamic &&\cite{yazdani2012new,yazdani2014mnafsa,yazdani2016novel}   \\
    
        \bottomrule
     \end{tabular}
   \end{adjustbox}
\end{table}

Yazdani et al.~\cite{yazdani2012new,yazdani2014mnafsa,yazdani2016novel} introduced a multi-population AFSA (mAFSA) model for optimization in dynamic environments where several additional components have been used to address the DOP challenges.
In all three versions of mAFSA, several AFSA sub-populations are responsible for covering multiple moving promising regions in the search space to improve the capability of the algorithm in reacting to the environmental changes and tracking the optimum position.
To improve the performance of each sub-population in mAFSA and increase their convergence speed, which is vital for tracking the optimum movement in each environment, several modifications have been incorporated into the to AFSA behaviors and parameters.
In the modified AFSA, $\delta$ is removed and $s$ is replaced by $v$, i.e., the step size and the visual length are unified.
In the preying behavior, the AFs are allowed to directly move to a better-found position; therefore, an AF may update its position $\mathfrak{t}$ times where it executes the preying behavior.
Besides that, in the following behavior, the best AF is used as the global attractor in each sub-population and is used to replace the neighbor AFs.
Although the modifications improve the exploitation ability of sub-populations significantly, their exploration ability is deteriorated.
This shortcoming is addressed by using the multi-population approach as well as some randomization operators performed by other components such as the exclusion~\cite{Blackwell2006Multiswarms} and initialization of new sub-populations.\par

\begin{figure}[tb!]
\centering
\includegraphics[width=0.55\linewidth]{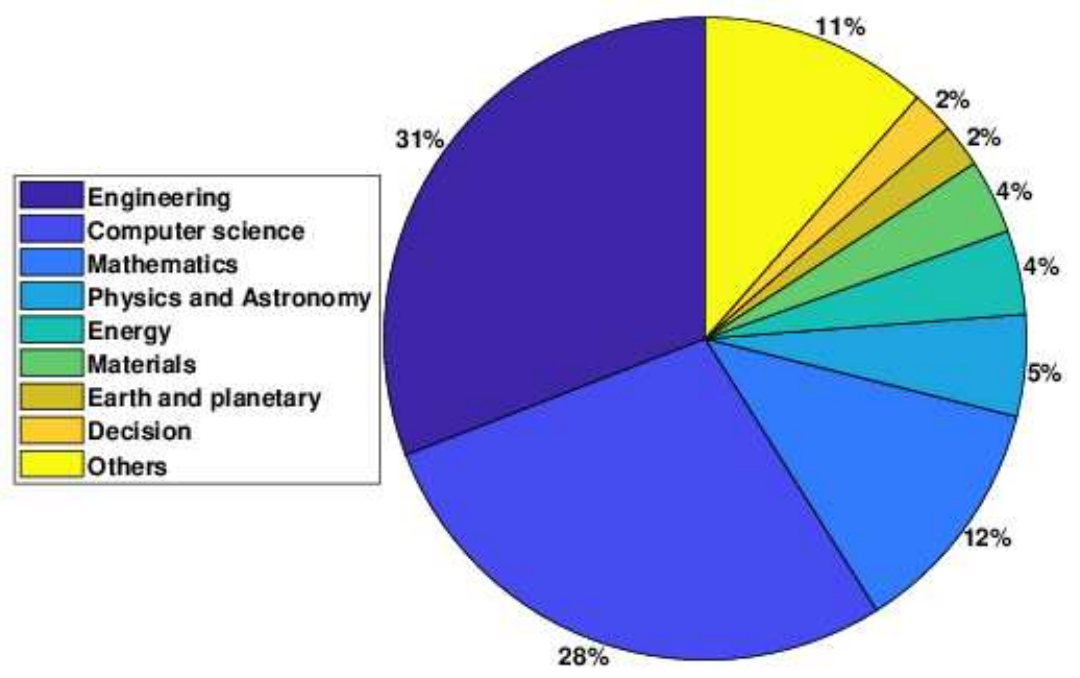}
\caption{\color{black}The AFSA application domains between 2004 and early 2021, as returned by conducting searches using the exact phrases of “Artificial Fish Swarm Algorithm”, “Artificial Fish Swarm Optimization”, and “Artificial Fish Schooling Algorithm” in title, abstract, or keywords of the retrieved articles. Source: {\color{black}https://www.scopus.com} }
\label{Fig:app}
\end{figure}

\section{Applications }
\label{Sec:app}
{\color{black}
Fig.~\ref{Fig:app} shows the application domains of AFSA and its variants from 2002 to early 2021. In addition, Table~\ref{Table:app} shows the applications of the reviewed articles. As can be seen (Fig.~\ref{Fig:app}), the AFSA and its variants have been applied to a wide range of application domains. Among them, ``engineering", ``computer science" and ``mathematics" are the most popular disciplines, with 31\%, 28\% and 12\%, respectively.\par 
}

{\color{black}
The reviewed articles (Table~\ref{Table:app}) indicate the specific applications within engineering, computer science, and mathematics disciplines.  In engineering, the AFSA and its variants are frequently used in the following areas: \textit{(i)} \textbf{sensor networks}, which include communication optics system, wireless sensor network (WSN) coverage optimization, routing protocol in WSN, scheduling of multi-access edge networks, holes recovery and water quality monitoring using sensor networks; \textit{(ii)} \textbf{control systems}, which include Proportional-Integral and Derivative control of the electronic stability program (ESP), field calibration, re-calibration of fiber optic gyroscope, tuned mass damper (TMD) optimization, underwater vehicle control, active disturbance rejection controller, distributed control and green wave traffic control; \textit{(iii)} \textbf{power systems}, which include home energy management system, economic dispatching of electric power system, power allocation schemes; and \textit{(iv)} \textbf{miscellaneous problems}, which include heat transfer, localization, tracking and navigation tasks.}\par

{\color{black}
The most popular applications of AFSA-based algorithms in computer science include \textit{(i)} \textbf{image processing}, e.g., classification and segmentation; \textit{(ii)} \textbf{signal processing}, e.g., classification and approximation; \textit{(iii)} \textbf{forecasting}, e.g., composite production, thermal error optimization, risk probability prediction, network traffic forecast, time series forecasting, annual water consumption prediction, and privacy services.  On the other hand, many \textbf{benchmark problems in mathematics} e.g., classification, clustering and regression, have been used to evaluate the effectiveness of AFSA-based algorithms.\par

In addition to abovementioned application domains, ``physics and astronomy", ``energy", ``materials", ``earth and planetary" and ``decision" are the subsequent application domains with 5\%, 4\%, 4\%, 2\% and 2\% coverage, respectively.  
The remaining 11\% are covered by ``Others", where each domain constitutes lower than 1\%, namely ``social sciences", ``environmental science", ``biochemistry, genetics and molecular biology", ``agricultural and biological sciences", ``chemistry", ``multidisciplinary", ``chemical engineering", ``neuroscience", ``business, management and accounting", ``medicine", ``arts and humanities", ``economics, econometrics and finance", ``immunology and microbiology", ``pharmacology, toxicology and pharmaceutics", ``psychology" and ``health professions", as indicated in Fig.~\ref{Fig:app}.
}

\begin{table}[tb]
\centering
\caption{\label{Table:app}\color{black} Applications of reviewed articles. }
    \begin{tabular}{l c c}
    \toprule
    Application &  Study \\ 
      \midrule
   \multirow{2}{*} {Benchmark} & \cite{yazdani2016novel,yazdani2013new,zhu2017adaptive,shao2017novel,zhou2019guidance,zhuang2019mechanical,gao2020twin,xi2019adaptive,peng2018modification,tan2019normative,zhu2020random,liu2019improved,wang2016afsaocp,huang2015log,li2015quantum,wei2019optimization,upadhyay2021periodic}\\
    & \cite{yaseen2018optimization,hajisalem2018hybrid,he2016hybrid,zheng2020hybrid,fang2014hybrid,li2016hybrid,serapiao2016combining,wang2015blind,sathya2017hybrid,xu2019integrated,yazdani2012new,yazdani2014mnafsa,yazdani2010new,yazdani2010CA}\\
   Sensor networks & \cite{mechta2017prolonging,feng2020analysis,yan2020wireless,qin2018adaptive,dawei2015wireless,cao2017modified,li2021computation,huang2021layout,zhou2021chaotic,Krishnaraj20212artificial,gorgich2021proposing} \\
   Control systems & \cite{Gao2014optimal,chen2016improved,li2018fault,gao2015novel,kang2019optimization,cheng2018research,he2019multi,jiang2017application,ma2019green} \\
   Image processing & \cite{lei2018image,liu2016complexity,zhang2017fish,el2015image,ma2015hybrid,yazdani2011color,goluguri2021rice}    \\
   Signal processing & \cite{fang2017estimation,zhang2019adaptive}   \\
   Forecasting & \cite{li2019risk,yan2020application,cheng2017parameter,yuan2020asafsa,xian2018novel,guo2016application,liu2016study,zhang2013identifying}   \\
   Localization, tracking & \multirow{2}{*}  {\cite{Du2018Precision,mao2018comprehensive,zhang2016robot,gao2015triaxial,fei2017application,sun2017application}}\\
   and navigation & \\
   Power systems & \cite{zhou2018artificial,Talha2018Energy,chen2016improved,yuan2019study,zhang2017adaptive,hua2021misalignment} \\
   \multirow{2}{*} {Others} & Circuit design~\cite{maji2018optimal}, motion estimation~\cite{Fei2014Motion}, water supply\\
   &  network~\cite{liu2019modified},
   privacy service~\cite{jia2020privacy}, heat transfer~\cite{wang2018estimation}\\
        \bottomrule
     \end{tabular}
\end{table}

{\color{black}
\section{Empirical Study and Potential Future Research Directions}
\label{Sec:future}
In this section, we first conduct an empirical study to compare the original AFSA and its variants with other population-based algorithms, then discuss potential future research directions.

\subsection{Empirical Study}
\label{Sec:emp}
In this section, the original AFSA and four recently published AFSA-based algorithms, namely NAFSA~\cite{tan2019normative}, CIAFSA~\cite{mao2018comprehensive}, PSO-FSA~\cite{duan2016improved} and PSOEM-FSA~\cite{duan2016improved}, are compared with PSO, DE~\cite{storn1997differential}, FA~\cite{nand2021stepping}, GWO~\cite{mirjalili2014grey} and CSA~\cite{askarzadeh2014anovel}.
{\color{black} The main reason of selecting NAFSA, CIAFSA, PSO-FSA and PSOEMFS, is their better performance as compared with other AFSA-variants reported in the literature.  }
Among the comparison methods, PSO is the most popular SI-based algorithms (Fig.~\ref{Fig:com}) used for solving a variety of optimization problems. PSO guides each solution (particle) by the best individual position (personal best) as well as the best population position (global best) to move toward a better position in the search space. It updates the velocity of each particle based on its previous velocity, personal best and global best positions. In PSO, the inertia weight ($w$) plays an important role, where it specifies how the particle’s previous velocity is modified in the next iteration. Other parameters of PSO are $C_1$ and $C_2$, i.e., the cognitive and social parameters, respectively. 
DE is a simple yet effective evolutionary computation (EC)-based optimization algorithm. It continuously improves the individuals through an iterative process that consists of three operations, i.e., mutation, crossover and selection. These operations are performed in a sequence at each iteration to produce a better solution. The scaling rate ($f$), top $p$\% individuals, and crossover rate are three important parameters in DE that need to be fine-tuned.\par 

Besides that, FA, GWO and CSA are three recently proposed SI-based algorithms. The FA~\cite{yang2010nature} is a stochastic, nature-inspired algorithm that imitates the characteristics of fireflies. Fireflies use flashing lights to attract mating partners and warn potential predators. There are three governing rules in the FA: \textit{(i)} fireflies are unisex, \textit{(ii)} attractiveness is proportional to brightness, and \textit{(iii)} the brightness of fireflies are proportional to the value of the fitness function. In the FA, a firefly’s attractiveness plays an important role, which depends on the maximum attractiveness value ($\beta_0$), absorption coefficient ($\gamma$) and distance between two fireflies $r_{ij}$.\par

\begin{table}[tb]
\centering
\caption{\label{Table:param}{\color{black} Parameters of AFSA, PSO, DE, FA, GWO and CSA that are required to be optimized.} }
    \begin{tabular}{l c }
    \toprule
    Algorithm &  Parameters \\ 
      \midrule
     AFSA & Crowding factor ($\delta$), visual range ($v$) and step ($s$)\\
     \midrule
     PSO & Inertia weight ($w$), cognitive parameters $C_1$ and social parameter $C_2$\\
     \midrule
     DE & Scaling rate ($f$), top $p$\% individuals and crossover rate\\
     \midrule
     FA & Maximum attractiveness value ($\beta_0$), absorption coefficient ($\gamma$) and $r_{ij}$\\
     \midrule
     GWO & $a$ and $C$\\
     \midrule
     CSA & Flight length ($fl$)\\
             \bottomrule
     \end{tabular}
\end{table}

GWO~\cite{mirjalili2014grey} mimics the leadership hierarchy and hunting mechanism of grey wolves in nature. GWO employs four types of grey wolves, i.e., alpha, beta, delta, and omega, to simulate the leadership hierarchy. Alpha, beta and delta, which indicate the first, second and third best positions, respectively, are used to guide hunting (optimization), while the remaining solutions are considered as omega. GWO implements three steps for hunting: \textit{(i)} searching for prey, \textit{(i)} encircling prey, and \textit{(iii)} attacking prey.\par

CSA~\cite{askarzadeh2014anovel} is another SI-based algorithm inspired by the intelligent group behavior of crows. Crows live in a group and hide their food in secret locations. Crows can recall looks and warn other crows if they identify any unfriendly looks. In addition, they watch other crows’ food hiding places and steal them. If a crow suspects the presence of a thief (another crow), it moves (with an awareness probability, $AP$) to a location far from the hidden place of food to fool the thief. CSA is based on four principles of crows in the nature: \textit{(i)} they live in the form of a group, \textit{(ii)} they remember the location of their hiding spots, \textit{(iii)} they chase other crows to steal their food, and \textit{(iv)} they protect their storages with a probability. The searching capability of CSA depends on the flight length ($fl$). A high value of $fl$ increases the global search ability of the algorithm, and vice versa.\par

Table~\ref{Table:param} summaries the parameters of AFSA, PSO, DE, FA, GWO and CSA that are required to be optimized in order to produce the best results. \par

Five benchmark functions are employed for performance evaluation, as summarized in  Table~\ref{Table:bench}. 
Table~\ref{Table:resfun} shows the results (mean and standard deviation, $SD$) of the compared methods, where the results of AFSA, PSO and DE are derived in this study by implementing the codes, and those of others are extracted from the corresponding references.
Overall, the AFSA and its variants perform comparatively well with other population-based algorithms.
AFSA and its variants are able to approximately reach the global minimum for most of the benchmark problems. In addition, relatively small standard deviations $(SD)$ on most of the benchmark functions indicate that the AFSA and its variants are highly robust and precise.\par

\begin{table}[tb]
\centering
\caption{\label{Table:bench}{ \color{black} Details of benchmark functions.} }
    \begin{tabular}{l c l c c}
    \toprule
    No. &  Test function & Expression & Domain &  Dimension\\ 
      \midrule
      $f_1$ & Sphere & $\sum_{i=1}^n x_i^2$  & [-100, 100]& 30\\ 
      \cmidrule{2-5}
      \multirow{2}{*} {$f_2$} & \multirow{2}{*} {Ackley} & $-20\exp (-0.2\sqrt {\frac{1}{n}\sum_{i=1}^n x_i^2)}$ & \multirow{2}{*} {[-32,32]} &\multirow{2}{*} {30}\\  
      &&~~~~~~~$-\exp(\frac{1}{n}\sum_{i=1}^n\cos(2\pi x_i)+20+e)$&  &  \\
      \cmidrule{2-5}
      $f_3$ & Alipine & $\sum_{i=1}^n|x_i\sin(x_i)+0.1x_i|$ & [-1,1] & 30\\ 
      \cmidrule{2-5}
      $f_4$ & Griewank & $\frac{1}{4000}\sum_{i=1}^n x_i^2-\prod_{i=1}^n \cos (\frac{x_i^2}{\sqrt{i}})+1$ & [-600,600] & 30 \\ 
      \cmidrule{2-5}
      $f_5$ & Quadric & $\sum_{i=1}^n\big(\sum_{j-1}^i x_j\big)^2$& [-100,100] & 30\\ 
           \bottomrule
     \end{tabular}
\end{table}
}
\subsection{ Potential Future Research Directions}
\label{Sec:futuree}

Based on the current research status of the field, the following potential future research directions are discussed.\par

\subsubsection{Improving AFSA}
Despite a large body of work, the AFSA and its variants are normally computationally complex due to the computation of many Euclidean distances for determining the neighborhood AFs and obtaining the unit vectors to limit the movement lengths to $s$ in the update rules of different behaviors.  
In addition, the convergence speed of the AFSA and its variants is usually slow, since the movement length is limited by $s$. 
Therefore, there are rooms for further improvement of the AFSA in terms of computational speed.\par

\subsubsection{Modifying AFSA for tackling different types of the optimization problems}

\begin{itemize}
\item\textit{Large-scale optimization Problems (LSOPs)}: 
The AFSA is yet to be comprehensively utilized for addressing LSOPs. 
LSOPs are high-dimensional problems (usually more than 100 dimensions) that pose the \emph{curse of dimensionality} or \emph{scalibility issues}~\cite{omidvar2014cooperative}. 
Many SI algorithms including the AFSA can easily be trapped in local optima in such problems. Considering the capability of the AFSA in maintaining diversity with the crowding factor and the individual local search of the AFs by performing the preying behavior, the AFSA has the potential to address the typical challenges of LSOPs. 
In this regard, effective formulae to ensure the scalability of the AFSA and its variants need to be derived.

\begin{table}[tb]
\centering
\caption{\label{Table:resfun} {\color{black} Performance comparison (mean and standard deviation, SD) of the AFSA and its variants with other population-based methods.  ``NA" indicates that the function is not used for evaluation with the corresponding method in the respective publication.}}
   \begin{adjustbox}{width=01.1\textwidth}
    \begin{tabular}{l c c| c c| c c| c c| c c}
    \toprule
    \multirow{2}{*} {Algorithm} &   \multicolumn{2}{c}{$f_1$} &  \multicolumn{2}{c}{$f_2$} &  \multicolumn{2}{c}{$f_3$} &   \multicolumn{2}{c}{$f_4$}&  \multicolumn{2}{c}{$f_5$}\\ 
      \cmidrule{2-11}
       & mean & SD& mean & SD& mean & SD& mean & SD& mean & SD\\
      \midrule
      AFSA & 1.21e-07 &  1.84e-07 &  3.09e-02 &  1.94e-02 & 4.14e-06 & 3.62e-06 & 7.11e-6  &7.57e-05 &  5.44e-05 & 1.29e-04 \\
      \midrule
      NAFSA~\cite{tan2019normative} &\hl{1.14e-52} &1.64e-52 &\hl{3.73e-15} &1.50e-15 & \hl{3.56e-26} &9.75e-08& 3.49e-02 & 1.68e-02 & 4.75e-09 & 9.13e-09 \\
      \midrule
      CIAFSA~\cite{mao2018comprehensive}& 6.51e-03 &2.59e-03 & 3.56e-04 &  3.95e-04 &  8.98e-06  & 1.18e-05 &  1.52e-02 & 3.47e-01 & 9.34e+00 &  2.11e+01\\
      \midrule
      PSO-FSA~\cite{duan2016improved} &1.12e-06 & 1.3107e-06 &2.32e-04 & 2.60e-04 & 1.27e-05  & 8.10e-06 &   1.44e-06 &2.47e-06 & 2.28e-04  &  2.90e-04\\
      \midrule
      PSOEM-FSA~\cite{duan2016improved}& 6.19e-08 & 1.11e-07 & 1.02e-04 & 1.11e-04 & 6.87e-06  & 5.39e-06 & 1.07e-01 &  1.84e-01 & 1.16e-04 & 2.63e-05 \\
      \midrule
      FA~\cite{nand2021stepping} & -2.2e1 & 1.13& -2.001e2 & 0 &NA & NA & \hl{4.44e-16} &0 & NA & NA \\
      \midrule
      CSA~\cite{askarzadeh2014anovel} & 6.9e-2 & 3.01e-2 &  4.468 & 0.676 & 1.816 &  0.280 & 0.066 &  201.307 & 201.307 & 43.371 \\
      \midrule
      GWO~\cite{mirjalili2014grey} & 6.59e-28 & 6.34e-05& 1.06e-13 & 7.7e-2 & 8.75e-06 & 1.95e-05 &  4.4e-3 & 6.65e-3 &   3.29e-6& 79.14 \\
      \midrule
      DE& 8.2e-14 & 5.9e-14 & 9.7e-08& 4.2e-08 & 2.29e-5 & 1.34e-5 & 4.66e-4& 1.21e-4&  \hl{6.8e-11} & 7.4e-11 \\
      \midrule
      PSO &  9.97e-09 & 1.40e-08 & 4.1e-4 & 9.10e-4 &  5.01e-4 & 5.16e-4 &  5.19e-06  & 2.32e-05 &  1.54e+03 & 4.91e+02 \\
           \bottomrule
     \end{tabular}
   \end{adjustbox}
\end{table}

 \item\textit{Multi-modal optimization problems (MMOPs)}:  
The AFSA  has not been adequately investigated for MMOPs.
It can be used to find multiple optimum solutions in multi-modal optimization problems.
In fact, by enhancing the adaptation of the crowding factor pertaining to the swarming behavior, the AFSA can maintain its overall diversity. 
Furthermore, by using the following behavior with distance-base neighborhoods, the population can be naturally divided into multiple species (sub-populations), which are attracted by different regions (peaks).
In addition to the following behavior, a local search capability can be incorporated, in order to allow the AFSA to find multiple peaks in complex multi-modal environments.

  \item\textit{Dynamic optimization problems (DOPs)}: 
  To date, few studies on modifying the AFSA for tackling DOPs are available in the literature. 
  In addressing DOPs, the AFSA and its variants can offer effective solutions. One important challenge of the DOPs is diversity loss in the population. 
  Indeed, algorithms whose populations collapse with respect to previous optimum solutions (due to the convergence issue) normally are ineffective in tracking the new optimum position after environmental changes. 
  One way to address diversity loss is to maintain the population diversity over time~\cite{lung2010evolutionary}. 
  Using the crowd factor, the AFSA can maintain its diversity to improve its capability of tracking the optimum movement. 
  Designing dynamic optimization algorithms with the AFSA as the core optimizer is a fruitful direction of future work.
\end{itemize}

\subsubsection{Analyzing the AFSA performance in different search spaces}
In the current literature, the AFSA performance pertaining to problems with different characteristics is yet to be fully investigated.  
As an example, the effectiveness of the AFSA with respect to rotated problems where the variable interaction structure is highly non-separable is not entirely clear. 
In addition, the AFSA performance on composition-based problems~\cite{liang2005novel} where the problem is partially separable or perhaps contains components with overlapping variables~\cite{omidvar2014cooperative} needs to be comprehensively studied.  
Besides that, the usefulness of the AFSA on problems with ill-conditioning and asymmetric properties requires further investigations. 
Therefore, a comprehensive analysis of the AFSA performance in different classes of optimization problems with different characteristics is an important future research direction.
Besides that, the usefulness of the AFSA on problems with ill-conditioning and asymmetric properties requires further investigations. Therefore, a comprehensive analysis of the AFSA performance in different classes of optimization problems with different characteristics is an important future research direction.

\subsubsection{Theoretical investigations}
{\color{black}
The AFSA has often been investigated empirically, and little attention has been given to the theoretical studies. Due to the importance of principled investigations in understanding the characteristics of the optimization capability of the AFSA and the associated advantages and disadvantages, additional detailed studies pertaining to theoretical aspects of the AFSA are necessary.  Specifically, a theoretical proof of convergence is vital to better understand the search behavior of the AFSA and its variants. This can be conducted by leveraging the stochastic process theory~\cite{jiang2007stochastic}. In addition, the Big-O notation analysis can be conducted to evaluate the runtime efficiency with respect to the growth in the input size, which is important for use in real-world environments. }
 
\section{Conclusions}
\label{Sec:con}
In this paper, a detailed review of the AFSA, which is a SI-based optimization algorithm inspired by the ecological behaviors of the fish swarm in nature, has been presented.  
We have focused on the AFSA and its variants that have been used to tackle continuous optimization problems.  
A taxonomy of the AFSA literature and a discussion with respect to the modifications and improvements of the AFSA parameters, procedures and sub-functions, hybrid formulations, multi-objective optimization, as well as dynamic AFSA models along with their applications have been provided. 
A discussion on future research trends has also been provided.  
Our further work will focus on providing a comprehensive review of the discrete AFSA models.

\begin{acknowledgements}
This work was supported in part by the National Natural Science Foundation of China under Grants 62176160, 61976141 and 61732011, in part by the Natural Science Foundation of Shenzhen (University Stability Support Program) under Grant 20200804193857002, and in part by the Interdisciplinary Innovation Team of Shenzhen University.
\end{acknowledgements}

%
%
\bibliographystyle{spmpsci}
\bibliography{mybib}

\begin{thebibliography}{100}
\providecommand{\url}[1]{{#1}}
\providecommand{\urlprefix}{URL }
\expandafter\ifx\csname urlstyle\endcsname\relax
  \providecommand{\doi}[1]{DOI~\discretionary{}{}{}#1}\else
  \providecommand{\doi}{DOI~\discretionary{}{}{}\begingroup
  \urlstyle{rm}\Url}\fi

\bibitem{rifaei2015deploying}
{Al-Rifaie}, M.M., {Aber}, A., {Hemanth}, D.J.: Deploying swarm intelligence in
  medical imaging identifying metastasis, micro-calcifications and brain image
  segmentation.
\newblock IET Systems Biology \textbf{9}(6), 234--244 (2015)

\bibitem{alkeshuosh2017using}
{Alkeshuosh}, A.H., {Moghadam}, M.Z., {Mansoori}, I.A., {Abdar}, M.: Using pso
  algorithm for producing best rules in diagnosis of heart disease.
\newblock In: International Conference on Computer and Applications (ICCA), pp.
  306--311 (2017)

\bibitem{askarzadeh2014anovel}
Askarzadeh, A.: A novel metaheuristic method for solving constrained
  engineering optimization problems: Crow search algorithm.
\newblock Computers \& Structures \textbf{169}, 1--12 (2016)

\bibitem{babaee2020fuzzy}
Babaee~Tirkolaee, E., Goli, A., Weber, G.W.: Fuzzy mathematical programming and
  self-adaptive artificial fish swarm algorithm for just-in-time energy-aware
  flow shop scheduling problem with outsourcing option.
\newblock IEEE Transactions on Fuzzy Systems \textbf{28}(11), 2772--2783 (2020)

\bibitem{bastos2008novel}
Bastos~Filho, C.J., de~Lima~Neto, F.B., Lins, A.J., Nascimento, A.I., Lima,
  M.P.: A novel search algorithm based on fish school behavior.
\newblock In: IEEE International Conference on Systems, Man and Cybernetics,
  pp. 2646--2651 (2008)

\bibitem{binghui2006random}
Binghui, Y., Xiaohui, Y., Jinwen, W., Xianzhang, Q.: A random perturbation
  particle swarm optimization algorithm.
\newblock Computer Engineering \textbf{32}(12), 189--190 (2006)

\bibitem{Blackwell2006Multiswarms}
{Blackwell}, T., {Branke}, J.: Multiswarms, exclusion, and anti-convergence in
  dynamic environments.
\newblock IEEE Transactions on Evolutionary Computation \textbf{10}(4),
  459--472 (2006)

\bibitem{blum2008swarm}
Blum, C., Li, X.: Swarm intelligence in optimization.
\newblock In: Swarm intelligence, pp. 43--85. Springer (2008)

\bibitem{cai2010artificial}
Cai, Y.: Artificial fish school algorithm applied in a combinatorial
  optimization problem.
\newblock International Journal of Intelligent Systems and Applications
  \textbf{2}(1), 37 (2010)

\bibitem{cao2017modified}
Cao, J., Zhao, X., Li, Z., Liu, W., Gu, H.: Modified artificial fish school
  algorithm for free space optical communication with sensor-less adaptive
  optics system.
\newblock Journal of the Korean Physical Society \textbf{71}(10), 636--646
  (2017)

\bibitem{chen2016improved}
Chen, L., Zhao, X.: An improved power control {AFSA} for minimum interference
  to primary users in cognitive radio networks.
\newblock Wireless Personal Communications \textbf{87}(1), 293--311 (2016)

\bibitem{chen2018application}
Chen, W., Feng, Y.Z., Jia, G.F., Zhao, H.T.: Application of artificial fish
  swarm algorithm for synchronous selection of wavelengths and spectral
  pretreatment methods in spectrometric analysis of beef adulteration.
\newblock Food Analytical Methods \textbf{11}(8), 2229--2236 (2018)

\bibitem{cheng2017parameter}
Cheng, M., Xiang, M.: Parameter estimation of a composite production function
  model based on improved artificial fish swarm algorithm and model
  application.
\newblock Communications in Statistics-Simulation and Computation
  \textbf{46}(10), 8218--8232 (2017)

\bibitem{cheng2009novel}
{Cheng}, Y., {Jiang}, M., {Yuan}, D.: Novel clustering algorithms based on
  improved artificial fish swarm algorithm.
\newblock In: IEEE International Conference on Fuzzy Systems and Knowledge
  Discovery, vol.~3, pp. 141--145 (2009)

\bibitem{cheng2018research}
Cheng, Z., Lu, Z.: Research on the {PID} control of the {ESP} system of tractor
  based on improved {AFSA} and improved {SA}.
\newblock Computers and Electronics in Agriculture \textbf{148}, 142--147
  (2018)

\bibitem{crepinsek2011analysis}
Crepinsek, M., Mernik, M., Liu, S.H.: Analysis of exploration and exploitation
  in evolutionary algorithms by ancestry trees.
\newblock International Journal of Innovative Computing and Applications
  \textbf{3}(1), 11--19 (2011)

\bibitem{dawei2015wireless}
DaWei, W., Changliang, W.: Wireless sensor networks coverage optimization based
  on improved {AFSA} algorithm.
\newblock International Journal of Future Generation Communication and
  Networking \textbf{8}(1), 99--108 (2015)

\bibitem{dorigo1997ant}
Dorigo, M., Gambardella, L.M.: Ant colony system: a cooperative learning
  approach to the traveling salesman problem.
\newblock IEEE Transactions on Evolutionary Computation \textbf{1}(1), 53--66
  (1997)

\bibitem{Du2018Precision}
{Du}, C., {Sun}, X., {Zhou}, J., {Dai}, Z., {Yin}, D.: Precision distribution
  method of navigation system based on improved artificial fish swarm
  algorithm.
\newblock In: 2018 10th International Conference on Intelligent Human-Machine
  Systems and Cybernetics (IHMSC), vol.~02, pp. 329--334 (2018)

\bibitem{duan2016improved}
Duan, Q., Mao, M., Duan, P., Hu, B.: An improved artificial fish swarm
  algorithm optimized by particle swarm optimization algorithm with extended
  memory.
\newblock Kybernetes \textbf{45}(2), 210--222 (2016)

\bibitem{duan2011simulation}
Duan, Q.C.: Simulation analysis of particle swarm optimization algorithm with
  extended memory.
\newblock Control and Decision \textbf{26}(7) (2011)

\bibitem{el2015image}
El-Said, S.A.: Image quantization using improved artificial fish swarm
  algorithm.
\newblock Soft Computing \textbf{19}(9), 2667--2679 (2015)

\bibitem{fang2014hybrid}
Fang, N., Zhou, J., Zhang, R., Liu, Y., Zhang, Y.: A hybrid of real coded
  genetic algorithm and artificial fish swarm algorithm for short-term optimal
  hydrothermal scheduling.
\newblock International Journal of Electrical Power \& Energy Systems
  \textbf{62}, 617--629 (2014)

\bibitem{fang2017estimation}
Fang, Z., Hu, L., Qin, L., Mao, K., Chen, W., Fu, X.: Estimation of ultrasonic
  signal onset for flow measurement.
\newblock Flow Measurement and Instrumentation \textbf{55}, 1--12 (2017)

\bibitem{farzi2009efficient}
Farzi, S.: Efficient job scheduling in grid computing with modified artificial
  fish swarm algorithm.
\newblock International Journal of Computer Theory and Engineering
  \textbf{1}(1), 13 (2009)

\bibitem{Fei2014Motion}
Fei, C., Zhang, P., Li, J.: Motion estimation based on artificial fish-swarm in
  {H}. 264/{AVC} coding.
\newblock WSEAS Transactions on Signal Processing \textbf{10}, 221--229 (2014)

\bibitem{fei2017application}
Fei, T., Zhang, L.: {Application of BFO-AFSA to location of distribution
  centre}.
\newblock Cluster Computing \textbf{20}(4), 3459--3474 (2017)

\bibitem{fei2021location}
Fei, T., Zhang, L., Zhang, X., Chen, Q., Liang, J.: Location selection strategy
  of distribution centers based on artificial fish swarm algorithm improved by
  bacterial colony chemotaxis.
\newblock Journal of Internet Technology \textbf{22}, 685--695 (2021)

\bibitem{feng2020analysis}
{Feng}, Y., {Zhao}, S., {Liu}, H.: Analysis of network coverage optimization
  based on feedback k-means clustering and artificial fish swarm algorithm.
\newblock IEEE Access \textbf{8}, 42864--42876 (2020)

\bibitem{Fernandes2009Fish}
Fernandes, E.M.G.P., Martins, T.F.M.C., Rocha, A.M.A.C.: Fish swarm intelligent
  algorithm for bound constrained global optimization.
\newblock In: International conference on computational and mathematical
  methods in science and engineering, pp. 1--12 (2009)

\bibitem{fister2013comprehensive}
Fister, I., Fister~Jr, I., Yang, X.S., Brest, J.: A comprehensive review of
  firefly algorithms.
\newblock Swarm and Evolutionary Computation \textbf{13}, 34--46 (2013)

\bibitem{Gao2014optimal}
Gao, Y., Guan, L., Wang, T.: Optimal artificial fish swarm algorithm for the
  field calibration on marine navigation.
\newblock Measurement \textbf{50}, 297 -- 304 (2014)

\bibitem{gao2015triaxial}
Gao, Y., Guan, L., Wang, T.: Triaxial accelerometer error coefficients
  identification with a novel artificial fish swarm algorithm.
\newblock Journal of Sensors \textbf{2015} (2015)

\bibitem{gao2015novel}
Gao, Y., Guan, L., Wang, T., Sun, Y.: A novel artificial fish swarm algorithm
  for recalibration of fiber optic gyroscope error parameters.
\newblock Sensors \textbf{15}(5), 10547--10568 (2015)

\bibitem{gao2020twin}
Gao, Y., Xie, L., Zhang, Z., Fan, Q.: {Twin support vector machine based on
  improved artificial fish swarm algorithm with application to flame
  recognition}.
\newblock Applied Intelligence  (2020)

\bibitem{gholami2018feature}
Gholami, J., Pourpanah, F., Wang, X.: Feature selection based on improved
  binary global harmony search for data classification.
\newblock Applied Soft Computing \textbf{93}, 106402 (2020)

\bibitem{goluguri2021rice}
Goluguri, N.R.R., Devi, K.S., Srinivasan, P.: Rice-net: an efficient artificial
  fish swarm optimization applied deep convolutional neural network model for
  identifying the oryza sativa diseases.
\newblock Neural Computing and Applications \textbf{33}(11), 5869--5884 (2021)

\bibitem{gorgich2021proposing}
Gorgich, S., Tabatabaei, S.: Proposing an energy-aware routing protocol by
  using fish swarm optimization algorithm in wsn (wireless sensor networks).
\newblock Wireless Personal Communications pp. 1--21 (2021)

\bibitem{guo2016application}
Guo, Q., Xu, R., Yang, T., He, L., Cheng, X., Li, Z., Yang, J.: Application of
  {GRAM} and {AFSACA-BPN} to thermal error optimization modeling of {CNC}
  machine tools.
\newblock The International Journal of Advanced Manufacturing Technology
  \textbf{83}(5-8), 995--1002 (2016)

\bibitem{hajisalem2018hybrid}
Hajisalem, V., Babaie, S.: {A hybrid intrusion detection system based on
  ABC-AFS algorithm for misuse and anomaly detection}.
\newblock Computer Networks \textbf{136}, 37--50 (2018)

\bibitem{he2019multi}
He, J., Jin, X., Xie, S., Cao, L., Lin, Y., Wang, N.: {Multi-body dynamics
  modeling and TMD optimization based on the improved AFSA for floating wind
  turbines}.
\newblock Renewable Energy \textbf{141}, 305--321 (2019)

\bibitem{he2016hybrid}
He, S., Belacel, N., Chan, A., Hamam, H., Bouslimani, Y.: A hybrid artificial
  fish swarm simulated annealing optimization algorithm for automatic
  identification of clusters.
\newblock International journal of information technology {\&} decision making
  \textbf{15}(05), 949--974 (2016)

\bibitem{he2021multi}
He, Y., Zhao, X., Guo, R., Gan, X.: Multi-resolution wavelet neural network
  learning algorithm based on artificial fish swarm algorithm.
\newblock In: The 2nd International Conference on Computing and Data Science,
  pp. 1--5 (2021)

\bibitem{hua2021misalignment}
Hua, Z., Xiao, Y., Cao, J.: Misalignment fault prediction of wind turbines
  based on improved artificial fish swarm algorithm.
\newblock Entropy \textbf{23}(6), 692 (2021)

\bibitem{huang2021layout}
Huang, J., Zeng, J., Bai, Y., Cheng, Z., Feng, Z., Qi, L., Liang, D.: Layout
  optimization of fiber bragg grating strain sensor network based on modified
  artificial fish swarm algorithm.
\newblock Optical Fiber Technology \textbf{65}, 102583 (2021)

\bibitem{huang2021optimization}
Huang, X., Xu, G., Xiao, F.: Optimization of a novel urban growth simulation
  model integrating an artificial fish swarm algorithm and cellular automata
  for a smart city.
\newblock Sustainability \textbf{13} (2021)

\bibitem{huang2015log}
Huang, Z., Chen, Y.: Log-linear model based behavior selection method for
  artificial fish swarm algorithm.
\newblock Computational Intelligence and Neuroscience \textbf{2015}, 10 (2015)

\bibitem{jia2020privacy}
Jia, B., Hao, L., Zhang, C., Huang, B.: A privacy-sensitive service selection
  method based on artificial fish swarm algorithm in the internet of things.
\newblock Mobile Networks and Applications pp. 1--9 (2020)

\bibitem{jia2020aparametric}
{Jia}, D., {Li}, Z., {Zhang}, C.: A parametric optimization oriented, afsa
  based random forest algorithm: Application to the detection of cervical
  epithelial cells.
\newblock IEEE Access \textbf{8}, 64891--64905 (2020)

\bibitem{jia2019animproved}
{Jia}, X., {Lu}, G.: An improved random taguchi's method based on swarm
  intelligence and dynamic reduced rate for electromagnetic optimization.
\newblock IEEE Antennas and Wireless Propagation Letters \textbf{18}(9),
  1878--1881 (2019)

\bibitem{jiang2017application}
Jiang, C., Wan, L., Sun, Y., Li, Y.: The application of {PSO-AFSA} method in
  parameter optimization for underactuated autonomous underwater vehicle
  control.
\newblock Mathematical Problems in Engineering \textbf{2017} (2017)

\bibitem{jiang2007stochastic}
Jiang, M., Luo, Y., Yang, S.: Stochastic convergence analysis and parameter
  selection of the standard particle swarm optimization algorithm.
\newblock Information Processing Letters \textbf{102}(1), 8--16 (2007)

\bibitem{kang2019optimization}
Kang, C., Wang, S., Ren, W., Lu, Y., Wang, B.: Optimization design and
  application of active disturbance rejection controller based on intelligent
  algorithm.
\newblock IEEE Access \textbf{7}, 59862--59870 (2019)

\bibitem{kanimozhi2021hybrid}
Kanimozhi, N., Singaravel, G.: Hybrid artificial fish particle swarm optimizer
  and kernel extreme learning machine for type-ii diabetes predictive model.
\newblock Medical \& Biological Engineering \& Computing \textbf{59}(4),
  841--867 (2021)

\bibitem{kennedy2010particle}
Kennedy, J.: Particle swarm optimization.
\newblock Encyclopedia of machine learning pp. 760--766 (2010)

\bibitem{koohestani2019integration}
{Koohestani}, A., {Abdar}, M., {Khosravi}, A., {Nahavandi}, S., {Koohestani},
  M.: Integration of ensemble and evolutionary machine learning algorithms for
  monitoring diver behavior using physiological signals.
\newblock IEEE Access \textbf{7}, 98971--98992 (2019)

\bibitem{Krishnaraj20212artificial}
Krishnaraj, N., Jayasankar, T., Kousik, N.V., Daniel, A.: 2 Artificial Fish
  Swarm Optimization Algorithm with Hill Climbing Based Clustering Technique
  for Throughput Maximization in Wireless Multimedia Sensor Network, pp.
  23--42.
\newblock River Publishers (2021)

\bibitem{kusakci2014adaptive}
Kusakci, A.O., Can, M.: An adaptive evolution strategy for constrained
  optimisation problems in engineering design.
\newblock International Journal of Bio-Inspired Computation \textbf{6}(3),
  175--191 (2014)

\bibitem{lei2018image}
Lei, X., Ouyang, H., Xu, L.: Image segmentation based on equivalent
  three-dimensional entropy method and artificial fish swarm optimization
  algorithm.
\newblock Optical Engineering \textbf{57}(10), 103106 (2018)

\bibitem{li2021diversity}
Li, C., Sun, J., Palade, V., Li, L.W.: Diversity collaboratively guided random
  drift particle swarm optimization.
\newblock International Journal of Machine Learning and Cybernetics pp. 1--22
  (2021)

\bibitem{li2019risk}
Li, H., Huang, Y., Tian, S.: Risk probability predictions for coal enterprise
  infrastructure projects in countries along the belt and road initiative.
\newblock International Journal of Industrial Ergonomics \textbf{69}, 110--117
  (2019)

\bibitem{li2015quantum}
Li, J., Zhao, S., Xu, Y.: Quantum-inspired artificial fish swarm algorithm
  based on the bloch sphere searching.
\newblock Quantum \textbf{4}(4), 06--18 (2015)

\bibitem{li2013artificial}
Li, S., Li, W., Sun, H.: Artificial fish swarm parallel algorithm based on
  multi-core cluster.
\newblock Journal of Computer Applications \textbf{33}(12), 3380--3384 (2013)

\bibitem{li2021computation}
Li, T., Yang, F., Zhang, D., Zhai, L.: Computation scheduling of multi-access
  edge networks based on the artificial fish swarm algorithm.
\newblock IEEE Access \textbf{9}, 74674--74683 (2021)

\bibitem{li2018fault}
Li, T.H., Xie, S.S., Liu, S.P., Xiao, L., Jia, W.Z., He, D.W.: A fault
  detection optimization method based on chaos adaptive artificial fish swarm
  algorithm on distributed control system.
\newblock Journal of Systems and Control Engineering \textbf{232}(9),
  1182--1193 (2018)

\bibitem{li2016hybrid}
Li, W., Bi, Y., Zhu, X., Yuan, C.a., Zhang, X.b.: Hybrid swarm intelligent
  parallel algorithm research based on multi-core clusters.
\newblock Microprocessors and Microsystems \textbf{47}, 151--160 (2016)

\bibitem{Li2002optimizing}
Li, X.L., Shao, Z.J., Qian, J.X.: {Optimizing method based on autonomous
  animats: fish-swarm algorithm}.
\newblock System Engineering Theory and Practice \textbf{22}(11), 32--38 (in
  Chinese) (2002)

\bibitem{liang2005novel}
Liang, J.J., Suganthan, P.N., Deb, K.: Novel composition test functions for
  numerical global optimization.
\newblock In: IEEE Swarm Intelligence Symposium, pp. 68--75. IEEE (2005)

\bibitem{lin2021kinematic}
Lin, M., Yuan, X., Lei, H., Ji, Z.: Kinematic analysis of tensegrity mechanisms
  based on improved artificial fish swarm algorithm with variable step size.
\newblock In: Journal of Physics: Conference Series, vol. 1903, p. 012071
  (2021)

\bibitem{liu2016complexity}
Liu, D., Zhao, D., Fu, Q., Wu, Q., Zhang, Y., Li, T., Imran, K.M., Abrar, F.M.:
  Complexity measurement of regional groundwater resources system using
  improved lempel-ziv complexity algorithm.
\newblock Arabian Journal of Geosciences \textbf{9}(20), 746 (2016)

\bibitem{liu2020solving}
Liu, Y., Feng, X., Yang, Y., Ruan, Z., Zhang, L., Li, K.: Solving urban
  electric transit network problem by integrating pareto artificial fish swarm
  algorithm and genetic algorithm.
\newblock Journal of Intelligent Transportation Systems pp. 1--28 (2020)

\bibitem{liu2019modified}
Liu, Y., Tao, Z., Yang, J., Mao, F.: The modified artificial fish swarm
  algorithm for least-cost planning of a regional water supply network problem.
\newblock Sustainability \textbf{11}(15), 4121 (2019)

\bibitem{liu2019improved}
Liu, Y., Wang, J., Shahbazzade, S.: The improved {AFSA} algorithm for the berth
  allocation and quay crane assignment problem.
\newblock Cluster Computing \textbf{22}(2), 3665--3672 (2019)

\bibitem{liu2016study}
Liu, Y., Wang, R.: Study on network traffic forecast model of {SVR} optimized
  by {GAFSA}.
\newblock Chaos, Solitons \& Fractals \textbf{89}, 153--159 (2016)

\bibitem{lung2010evolutionary}
Lung, R.I., Dumitrescu, D.: Evolutionary swarm cooperative optimization in
  dynamic environments.
\newblock Natural Computing \textbf{9}(1), 83--94 (2010)

\bibitem{ma2019green}
Ma, C., He, R.: Green wave traffic control system optimization based on
  adaptive genetic-artificial fish swarm algorithm.
\newblock Neural Computing and Applications \textbf{31}(7), 2073--2083 (2019)

\bibitem{ma2015hybrid}
Ma, L., Li, Y., Fan, S., Fan, R.: A hybrid method for image segmentation based
  on artificial fish swarm algorithm and fuzzy-means clustering.
\newblock Computational and Mathematical Methods in Medicine \textbf{2015}
  (2015)

\bibitem{maji2018optimal}
Maji, K.B., Kar, R., Mandal, D., Ghoshal, S.: Optimal design of low power high
  gain and high speed cmos circuits using fish swarm optimization algorithm.
\newblock International Journal of Machine Learning and Cybernetics
  \textbf{9}(5), 771--786 (2018)

\bibitem{mao2018comprehensive}
Mao, M., Duan, Q., Duan, P., Hu, B.: Comprehensive improvement of artificial
  fish swarm algorithm for global mppt in pv system under partial shading
  conditions.
\newblock Transactions of the Institute of Measurement and Control
  \textbf{40}(7), 2178--2199 (2018)

\bibitem{Mavrovouniotiz2017asurvey}
Mavrovouniotis, M., Li, C., Yang, S.: A survey of swarm intelligence for
  dynamic optimization: algorithms and applications.
\newblock Swarm and Evolutionary Computation \textbf{33}, 1 -- 17 (2017)

\bibitem{mechta2017prolonging}
Mechta, D., Harous, S.: Prolonging wsn lifetime using a new scheme for sink
  moving based on artificial fish swarm algorithm.
\newblock In: Proceedings of the Second International Conference on Advanced
  Wireless Information, Data, and Communication Technologies, pp. 1--5 (2017)

\bibitem{mirjalili2014grey}
Mirjalili, S., Mirjalili, S.M., Lewis, A.: Grey wolf optimizer.
\newblock Advances in Engineering Software \textbf{69}, 46--61 (2014)

\bibitem{nand2021stepping}
Nand, R., Sharma, B.N., Chaudhary, K.: Stepping ahead firefly algorithm and
  hybridization with evolution strategy for global optimization problems.
\newblock Applied Soft Computing \textbf{109}, 107517 (2021)

\bibitem{neshat2014artificial}
Neshat, M., Sepidnam, G., Sargolzaei, M., Toosi, A.N.: Artificial fish swarm
  algorithm: a survey of the state-of-the-art, hybridization, combinatorial and
  indicative applications.
\newblock Artificial intelligence review \textbf{42}(4), 965--997 (2014)

\bibitem{omidvar2014cooperative}
{Omidvar}, M.N., {Li}, X., {Mei}, Y., {Yao}, X.: Cooperative co-evolution with
  differential grouping for large scale optimization.
\newblock IEEE Transactions on Evolutionary Computation \textbf{18}(3),
  378--393 (2014)

\bibitem{pajouhi2018image}
{Pajouhi}, Z., {Roy}, K.: Image edge detection based on swarm intelligence
  using memristive networks.
\newblock IEEE Transactions on Computer-Aided Design of Integrated Circuits and
  Systems \textbf{37}(9), 1774--1787 (2018)

\bibitem{pavlyukevich2007levy}
Pavlyukevich, I.: L{\'e}vy flights, non-local search and simulated annealing.
\newblock Journal of Computational Physics \textbf{226}(2), 1830--1844 (2007)

\bibitem{peng2018modification}
Peng, Z., Dong, K., Yin, H., Bai, Y.: Modification of fish swarm algorithm
  based on levy flight and firefly behavior.
\newblock Computational Intelligence and Neuroscience \textbf{2018} (2018)

\bibitem{pourpanah2016ahybrid}
Pourpanah, F., Lim, C.P., Saleh, J.M.: A hybrid model of fuzzy artmap and
  genetic algorithm for data classification and rule extraction.
\newblock Expert Systems with Applications \textbf{49}, 74--85 (2016)

\bibitem{pourpanah2019ahybrid}
Pourpanah, F., Lim, C.P., Wang, X., Tan, C.J., Seera, M., Shi, Y.: A hybrid
  model of fuzzy min–max and brain storm optimization for feature selection
  and data classification.
\newblock Neurocomputing \textbf{333}, 440 -- 451 (2019)

\bibitem{pourpanah2019feature}
Pourpanah, F., Shi, Y., Lim, C.P., Hao, Q., Tan, C.J.: Feature selection based
  on brain storm optimization for data classification.
\newblock Applied Soft Computing \textbf{80}, 761 -- 775 (2019)

\bibitem{pourpanah2017aqlearning}
Pourpanah, F., Tan, C.J., Lim, C.P., Mohamad-Saleh, J.: A q-learning-based
  multi-agent system for data classification.
\newblock Applied Soft Computing \textbf{52}, 519--531 (2017)

\bibitem{pourpanah2019feature1}
{Pourpanah}, F., {Wang}, R., {Wang}, X.: Feature selection for data
  classification based on binary brain storm optimization.
\newblock In: IEEE International Conference on Cloud Computing and Intelligence
  Systems (CCIS), pp. 108--113 (2019)

\bibitem{pourpanah2019mbso}
{Pourpanah}, F., {Wang}, R., {Wang}, X., {Shi}, Y., {Yazdani}, D.: mbso: A
  multi-population brain storm optimization for multimodal dynamic optimization
  problems.
\newblock In: 2019 IEEE Symposium Series on Computational Intelligence (SSCI),
  pp. 673--679 (2019)

\bibitem{pourpanah2018anomaly}
{Pourpanah}, F., {Zhang}, B., {Ma}, R., {Hao}, Q.: Anomaly detection and
  condition monitoring of uav motors and propellers.
\newblock In: IEEE SENSORS, pp. 1--4 (2018)

\bibitem{pourpanah2018non}
Pourpanah, F., Zhang, B., Ma, R., Hao, Q.: Non-intrusive human motion
  recognition using distributed sparse sensors and the genetic algorithm based
  neural network.
\newblock In: 2018 IEEE SENSORS, pp. 1--4 (2018)

\bibitem{qin2018adaptive}
Qin, N., Xu, J.: An adaptive fish swarm-based mobile coverage in {WSNs}.
\newblock Wireless Communications and Mobile Computing \textbf{2018} (2018)

\bibitem{reynolds2004cultural}
Reynolds, R.G., Peng, B.: Cultural algorithms: modeling of how cultures learn
  to solve problems.
\newblock In: IEEE International Conference on Tools with Artificial
  Intelligence, pp. 166--172 (2004)

\bibitem{sathya2017hybrid}
Sathya, D.J., Geetha, K.: Hybrid {ANN} optimized artificial fish swarm
  algorithm based classifier for classification of suspicious lesions in breast
  {DCE-MRI}.
\newblock Polish Journal of Medical Physics and Engineering \textbf{23}(4),
  81--88 (2017)

\bibitem{serapiao2016combining}
Serapi{\~a}o, A.B., Corr{\^e}a, G.S., Gon{\c{c}}alves, F.B., Carvalho, V.O.:
  Combining {K}-means and {K}-harmonic with fish school search algorithm for
  data clustering task on graphics processing units.
\newblock Applied Soft Computing \textbf{41}, 290--304 (2016)

\bibitem{shao2017novel}
Shao, H., Jiang, H., Zhao, H., Wang, F.: A novel deep autoencoder feature
  learning method for rotating machinery fault diagnosis.
\newblock Mechanical Systems and Signal Processing \textbf{95}, 187--204 (2017)

\bibitem{shi2011brain}
Shi, Y.: Brain storm optimization algorithm.
\newblock In: International Conference in Swarm Intelligence, pp. 303--309
  (2011)

\bibitem{shi1998modified}
Shi, Y., Eberhart, R.: A modified particle swarm optimizer.
\newblock In: IEEE International Conference on Evolutionary Computation
  Proceedings, pp. 69--73 (1998)

\bibitem{storn1997differential}
Storn, R., Price, K.: Differential evolution--a simple and efficient heuristic
  for global optimization over continuous spaces.
\newblock Journal of global optimization \textbf{11}(4), 341--359 (1997)

\bibitem{sun2017application}
Sun, T., Zhang, H., Liu, S., Cao, Y.: Application of an artificial fish swarm
  algorithm in solving multiobjective trajectory optimization problems.
\newblock Chemistry and Technology of Fuels and Oils \textbf{53}(4), 541--547
  (2017)

\bibitem{Talha2018Energy}
Talha, M., Saeed, M.S., Mohiuddin, G., Ahmad, M., Nazar, M.J., Javaid, N.:
  Energy optimization in home energy management system using artificial fish
  swarm algorithm and genetic algorithm.
\newblock In: International Conference on Intelligent Networking and
  Collaborative Systems, pp. 203--213 (2018)

\bibitem{tan2019normative}
Tan, W.H., Mohamad-Saleh, J.: Normative fish swarm algorithm {(NFSA)} for
  optimization.
\newblock Soft Computing pp. 1--17 (2019)

\bibitem{upadhyay2021periodic}
Upadhyay, P., Pandey, M.K., Kohli, N.: Periodic pattern mining from
  spatio-temporal database using novel global pollination artificial fish swarm
  optimizer-based clustering and modified fp tree.
\newblock Soft Computing \textbf{25}(6), 4327--4344 (2021)

\bibitem{wang2015blind}
Wang, H., Guo, Y.: A blind equalization algorithm based on global artificial
  fish swarm and genetic optimization {DNA} encoding sequences.
\newblock In: Industrial Informatics and Computer Engineering Conference, pp.
  131--134 (2015)

\bibitem{wang2016afsaocp}
Wang, H.b., Fan, C.C., Tu, X.y.: {AFSAOCP}: A novel artificial fish swarm
  optimization algorithm aided by ocean current power.
\newblock Applied Intelligence \textbf{45}(4), 992--1007 (2016)

\bibitem{wang2018estimation}
Wang, X., Li, H., Li, Z.: Estimation of interfacial heat transfer coefficient
  in inverse heat conduction problems based on artificial fish swarm algorithm.
\newblock Heat and Mass Transfer \textbf{54}(10), 3151--3162 (2018)

\bibitem{wei2019optimization}
Wei, P., Li, Y., Zhang, Z., Hu, T., Li, Z., Liu, D.: An optimization method for
  intrusion detection classification model based on deep belief network.
\newblock IEEE Access \textbf{7}, 87593--87605 (2019)

\bibitem{xi2019adaptive}
Xi, L., Zhang, F.: An adaptive artificial-fish-swarm-inspired fuzzy c-means
  algorithm.
\newblock Neural Computing and Applications pp. 1--9 (2019)

\bibitem{xian2018novel}
Xian, S., Zhang, J., Xiao, Y., Pang, J.: A novel fuzzy time series forecasting
  method based on the improved artificial fish swarm optimization algorithm.
\newblock Soft Computing \textbf{22}(12), 3907--3917 (2018)

\bibitem{xian2021early}
Xian, Z., Yang, H.: An early warning model for the stuck-in medical drilling
  process based on the artificial fish swarm algorithm and svm.
\newblock Distributed and Parallel Databases pp. 1--18 (2021)

\bibitem{xu2019integrated}
Xu, H., Zhao, Y., Ye, C., Lin, F.: Integrated optimization for mechanical
  elastic wheel and suspension based on an improved artificial fish swarm
  algorithm.
\newblock Advances in Engineering Software \textbf{137}, 102722 (2019)

\bibitem{yan2020wireless}
Yan, L., He, Y., Huangfu, Z.: {A Fish Swarm Inspired Holes Recovery Algorithm
  for Wireless Sensor Networks}.
\newblock International Journal of Wireless Information Networks
  \textbf{27}(1), 89--101 (2020)

\bibitem{yan2020application}
Yan, W., Li, M., Pan, X., Wu, G., Liu, L.: Application of support vector
  regression cooperated with modified artificial fish swarm algorithm for wind
  tunnel performance prediction of automotive radiators.
\newblock Applied Thermal Engineering \textbf{164}, 114543 (2020)

\bibitem{yan2020anovel}
{Yan}, W., {Li}, M., {Zhong}, Y., {Qu}, C., {Li}, G.: A novel k-mpso clustering
  algorithm for the construction of typical driving cycles.
\newblock IEEE Access \textbf{8}, 64028--64036 (2020)

\bibitem{yang2010nature}
Yang, X.S.: Nature-inspired metaheuristic algorithms.
\newblock Luniver press (2010)

\bibitem{yang2010new}
Yang, X.S.: A New Metaheuristic Bat-Inspired Algorithm, pp. 65--74.
\newblock Springer (2010)

\bibitem{yang2009cuckoo}
Yang, X.S., Deb, S.: Cuckoo search via l{\'e}vy flights.
\newblock In: World congress on nature \& biologically inspired computing
  (NaBIC), pp. 210--214 (2009)

\bibitem{yaseen2018optimization}
Yaseen, Z.M., Karami, H., Ehteram, M., Mohd, N.S., Mousavi, S.F., Hin, L.S.,
  Kisi, O., Farzin, S., Kim, S., El-Shafie, A.: Optimization of reservoir
  operation using new hybrid algorithm.
\newblock Journal of Civil Engineering \textbf{22}(11), 4668--4680 (2018)

\bibitem{yazdani2012new}
Yazdani, D., Akbarzadeh-Totonchi, M.R., Nasiri, B., Meybodi, M.R.: A new
  artificial fish swarm algorithm for dynamic optimization problems.
\newblock In: EEE Congress on Evolutionary Computation, pp. 1--8. IEEE (2012)

\bibitem{yazdani2010CA}
Yazdani, D., Golyari, S., Meybodi, M.R.: A new hybrid algorithm for
  optimization based on artificial fish swarm algorithm and cellular learning
  automata.
\newblock In: International Symposium on Telecommunications, pp. 932--937. IEEE
  (2010)

\bibitem{yazdani2010new}
Yazdani, D., Golyari, S., Meybodi, M.R.: A new hybrid approach for data
  clustering.
\newblock In: International Symposium on Telecommunications, pp. 914--919. IEEE
  (2010)

\bibitem{yazdani2011color}
Yazdani, D., Nabizadeh, H., Kosari, E.M., Toosi, A.N.: Color quantization using
  modified artificial fish swarm algorithm.
\newblock In: Australasian Joint Conference on Artificial Intelligence, pp.
  382--391. Springer (2011)

\bibitem{yazdani2014mnafsa}
Yazdani, D., Nasiri, B., Sepas-Moghaddam, A., Meybodi, M., Akbarzadeh-Totonchi,
  M.: mnafsa: A novel approach for optimization in dynamic environments with
  global changes.
\newblock Swarm and Evolutionary Computation \textbf{18}, 38--53 (2014)

\bibitem{yazdani2013new}
Yazdani, D., Saman, B., Sepas-Moghaddam, A., Mohammad-Kazemi, F., Meybodi,
  M.R.: A new algorithm based on improved artificial fish swarm algorithm for
  data clustering.
\newblock International Journal of Artificial Intelligence \textbf{11}(13),
  1--29 (2013)

\bibitem{yazdani2016novel}
Yazdani, D., Sepas-Moghaddam, A., Dehban, A., Horta, N.: A novel approach for
  optimization in dynamic environments based on modified artificial fish swarm
  algorithm.
\newblock International Journal of Computational Intelligence and Applications
  \textbf{15}(02), 1650010 (2016)

\bibitem{yuan2019study}
Yuan, G., Yang, W.: Study on optimization of economic dispatching of electric
  power system based on hybrid intelligent algorithms {(PSO and AFSA)}.
\newblock Energy \textbf{183}, 926--935 (2019)

\bibitem{yuan2020asafsa}
Yuan, Y., Li, Q., Yuan, X., Luo, X., Liu, S.: {A SAFSA- and Metabolism-Based
  Nonlinear Grey Bernoulli Model for Annual Water Consumption Prediction}.
\newblock Iranian Journal of Science and Technology, Transactions of Civil
  Engineering \textbf{44}(2), 755--765 (2020)

\bibitem{zhang2017fish}
Zhang, F.s., Li, S.w., Hu, Z.g., Du, Z.: Fish swarm window selection algorithm
  based on cell microscopic automatic focus.
\newblock Cluster Computing \textbf{20}(1), 485--495 (2017)

\bibitem{zhang2021research}
Zhang, L., Fu, M., Fei, T.: Research on location of cold chain logistics
  distribution center with low carbon in beijing-tianjin-hebei area on the
  basis of rna-artificial fish swarm algorithm.
\newblock In: Journal of Physics: Conference Series, vol. 1861, p. 012005
  (2021)

\bibitem{zhang2021improved}
Zhang, L., Fu, M., Li, H., Liu, T.: Improved artificial bee colony algorithm
  based on damping motion and artificial fish swarm algorithm.
\newblock In: Journal of Physics: Conference Series, vol. 1903, p. 012038
  (2021)

\bibitem{zhang2017adaptive}
Zhang, S., Zhao, X., Liang, C., Ding, X.: {Adaptive power allocation schemes
  based on IAFS algorithm for OFDM-based cognitive radio systems}.
\newblock International Journal of Electronics \textbf{104}(1), 1--15 (2017)

\bibitem{zhang2021parameter}
Zhang, X., Lian, L., Zhu, F.: Parameter fitting of variogram based on hybrid
  algorithm of particle swarm and artificial fish swarm.
\newblock Future Generation Computer Systems \textbf{116}, 265--274 (2021)

\bibitem{zhang2013identifying}
Zhang, X., Wang, J., Yang, A., Yan, C., Zhu, F., Zhao, Z., Cao, Z.: Identifying
  interacting genetic variations by fish-swarm logic regression.
\newblock BioMed Research International \textbf{2013} (2013)

\bibitem{zhang2016robot}
Zhang, Y., Guan, G., Pu, X.: The robot path planning based on improved
  artificial fish swarm algorithm.
\newblock Mathematical Problems in Engineering \textbf{2016} (2016)

\bibitem{zhang2019adaptive}
Zhang, Z., Ma, J.: Adaptive parameter-tuning stochastic resonance based on
  {SVD} and its application in weak {IF} digital signal enhancement.
\newblock Journal on Advances in Signal Processing \textbf{2019}(1), 1--24
  (2019)

\bibitem{zhang2017pareto}
Zhang, Z., Wang, K., Zhu, L., Wang, Y.: A pareto improved artificial fish swarm
  algorithm for solving a multi-objective fuzzy disassembly line balancing
  problem.
\newblock Expert Systems with Applications \textbf{86}, 165--176 (2017)

\bibitem{zheng2020hybrid}
Zheng, R., Feng, Z., Shi, J., Jiang, S., Tan, L.: Hybrid bacterial forging
  optimization based on artificial fish swarm algorithm and gaussian
  disturbance.
\newblock In: Bio-inspired Computing: Theories and Applications, pp. 124--134
  (2020)

\bibitem{zhou2018artificial}
Zhou, G., Li, Y., He, Y.C., Wang, X., Yu, M.: Artificial fish swarm based power
  allocation algorithm for mimo-ofdm relay underwater acoustic communication.
\newblock IET Communications \textbf{12}(9), 1079--1085 (2018)

\bibitem{zhou2021chaotic}
Zhou, J., Qi, G., Liu, C.: A chaotic parallel artificial fish swarm algorithm
  for water quality monitoring sensor networks 3d coverage optimization.
\newblock Journal of Sensors \textbf{2021} (2021)

\bibitem{zhou2019guidance}
Zhou, X., Wang, Z., Li, D., Zhou, H., Qin, Y., Wang, J.: Guidance systematic
  error separation for mobile launch vehicles using artificial fish swarm
  algorithm.
\newblock IEEE Access \textbf{7}, 31422--31434 (2019)

\bibitem{zhu2017adaptive}
Zhu, J., Wang, C., Hu, Z., Kong, F., Liu, X.: Adaptive variational mode
  decomposition based on artificial fish swarm algorithm for fault diagnosis of
  rolling bearings.
\newblock Proceedings of the Institution of Mechanical Engineers, Part C:
  Journal of Mechanical Engineering Science \textbf{231}(4), 635--654 (2017)

\bibitem{zhu2020random}
Zhu, Y., XU, W., Luo, G., Wang, H., Yang, J., Lu, W.: Random forest enhancement
  using improved artificial fish swarm for the medial knee contact force
  prediction.
\newblock Artificial Intelligence in Medicine \textbf{103}, 101811 (2020)

\bibitem{zhuang2019mechanical}
Zhuang, D., Ma, K., Tang, C., Liang, Z., Wang, K., Wang, Z.: Mechanical
  parameter inversion in tunnel engineering using support vector regression
  optimized by multi-strategy artificial fish swarm algorithm.
\newblock Tunnelling and Underground Space Technology \textbf{83}, 425--436
  (2019)

\bibitem{zomorodi2019hybrid}
Zomorodi-moghadam, M., Abdar, M., Davarzani, Z., Zhou, X., P{\l}awiak, P.,
  Acharya, U.R.: Hybrid particle swarm optimization for rule discovery in the
  diagnosis of coronary artery disease.
\newblock Expert Systems p. e12485 (2019)

\end{thebibliography}


\end{document}